\documentclass[10pt,twocolumn,letterpaper]{article}

\usepackage{iccv}
\usepackage{times}
\usepackage{epsfig}
\usepackage{graphicx}
\usepackage{amsmath}
\usepackage{amssymb}
\usepackage{multirow}
\usepackage{subfigure}
\usepackage{graphicx}
\usepackage[usenames,dvipsnames]{color}
\usepackage{colortbl}
\usepackage{float}
\definecolor{mygray}{gray}{.9}

\usepackage[pagebackref=true,breaklinks=true,letterpaper=true,colorlinks,bookmarks=false]{hyperref}

\iccvfinalcopy 

\def\iccvPaperID{1381} 

\ificcvfinal\fi
\begin{document}

\title{Disentangled Image Matting}

\author{Shaofan Cai$^{1*}$, Xiaoshuai Zhang$^{1,2}$\thanks{Equal Contributors.This work is supported by The National Key Research and Development Program of China (2018YFC0831700).},  Haoqiang Fan$^1$, Haibin Huang$^1$,\\
 Jiangyu Liu$^1$, Jiaming Liu$^1$, Jiaying Liu$^2$, Jue Wang$^1$, and Jian Sun$^1$\\
$^1$Megvii Technology\\
$^2$Institute of Computer Science and Technology, Peking University\\
{\tt\small \{caishaofan, fhq, huanghaibin, liujiangyu, liujiaming, wangjue, sunjian\}@megvii.com}\\
{\tt\small \{jet, liujiaying\}@pku.edu.cn}
}

\newcommand{\jet}[1]{\textcolor{red}{zxs: #1}}

\maketitle

\begin{abstract}
Most previous image matting methods require a roughly-specificed trimap as input, and estimate fractional alpha values for all pixels that are in the unknown region of the trimap. In this paper, we argue that directly estimating the alpha matte from a coarse trimap is a major limitation of previous methods, as this practice tries to address two difficult and inherently different problems at the same time: identifying true blending pixels inside the trimap region, and estimate accurate alpha values for them.  
We propose AdaMatting, a new end-to-end matting framework that disentangles this problem into two sub-tasks: trimap adaptation and alpha estimation. Trimap adaptation is a pixel-wise classification problem that infers the global structure of the input image by identifying definite foreground, background, and semi-transparent image regions. Alpha estimation is a regression problem that calculates the opacity value of each blended pixel. Our method separately handles these two sub-tasks within a single deep convolutional neural network (CNN). Extensive experiments show that AdaMatting has additional structure awareness and trimap fault-tolerance. Our method achieves the state-of-the-art performance on Adobe Composition-1k dataset both qualitatively and quantitatively. It is also the current best-performing method on the \url{alphamatting.com} online evaluation for all commonly-used metrics.

\end{abstract}

\vspace{-0.4cm}
\section{Introduction}

\begin{figure}
\centering
\subfigure[]{
\begin{minipage}[t]{0.245\linewidth}
\includegraphics[width=1\linewidth]{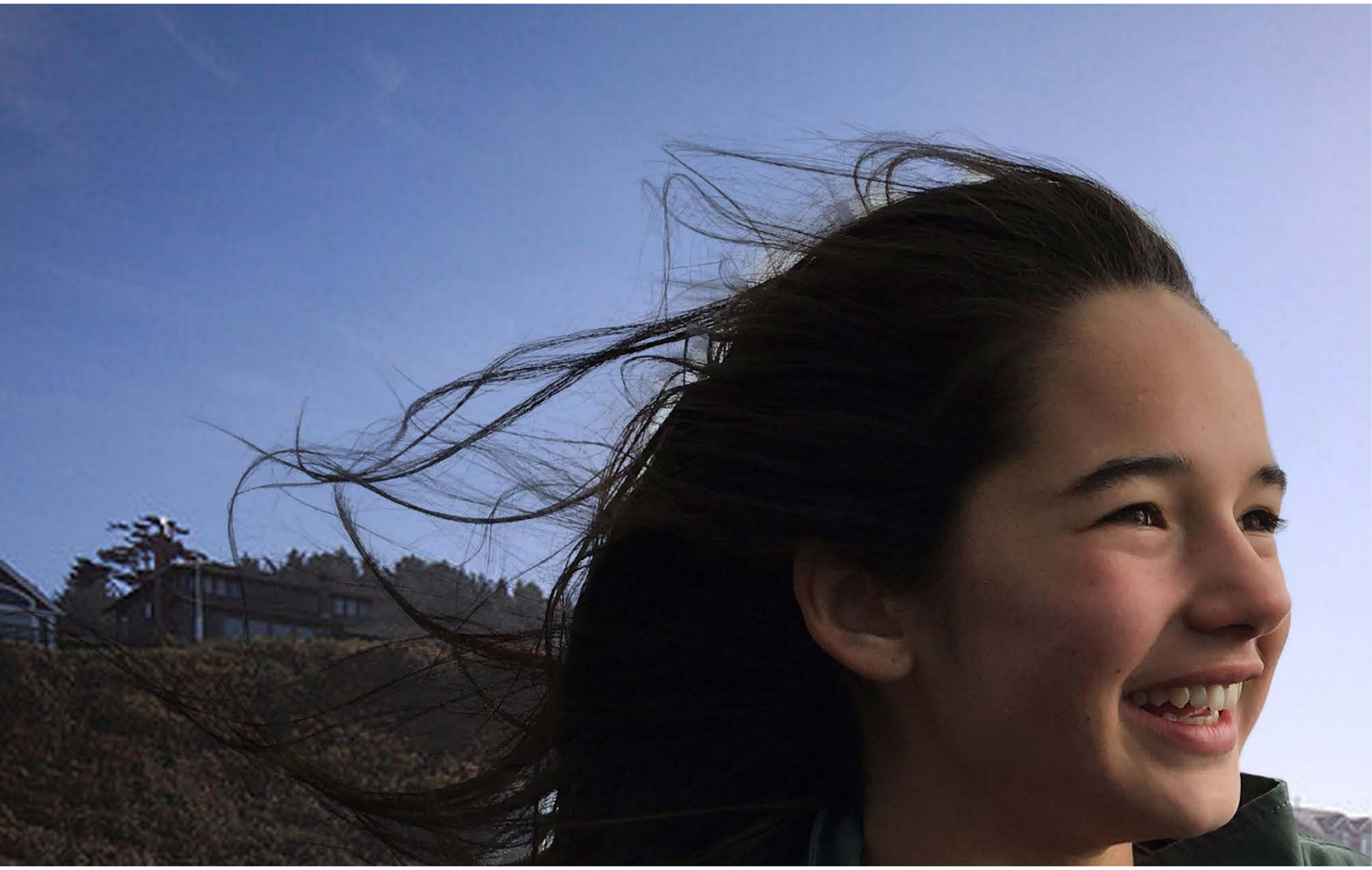}\vspace{-.08cm}
\includegraphics[width=1\linewidth]{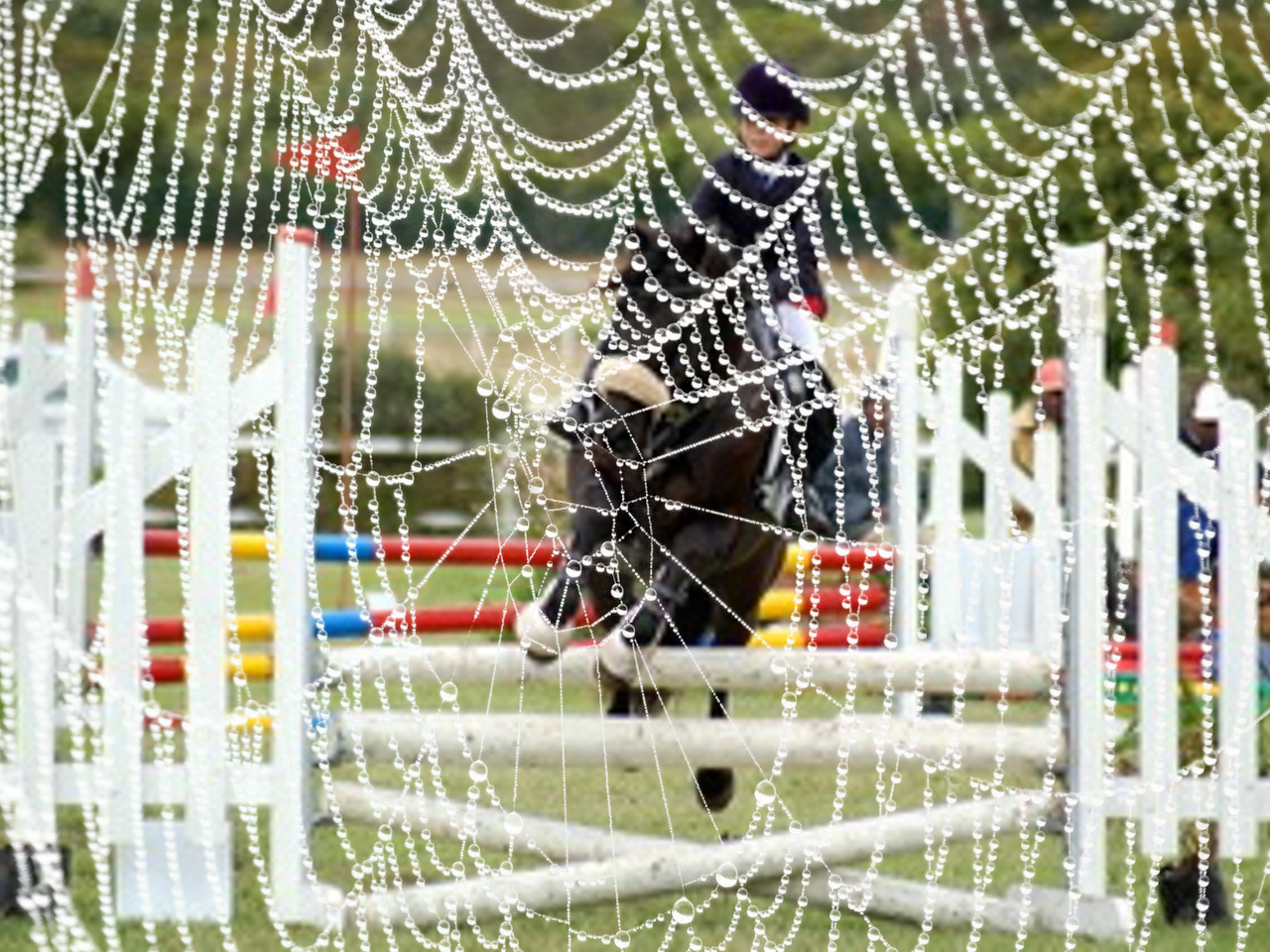}\vspace{-.25cm}
\label{img}
\end{minipage}}\hspace{-.15cm}
\subfigure[]{
\begin{minipage}[t]{0.245\linewidth} 
\includegraphics[width=1\linewidth]{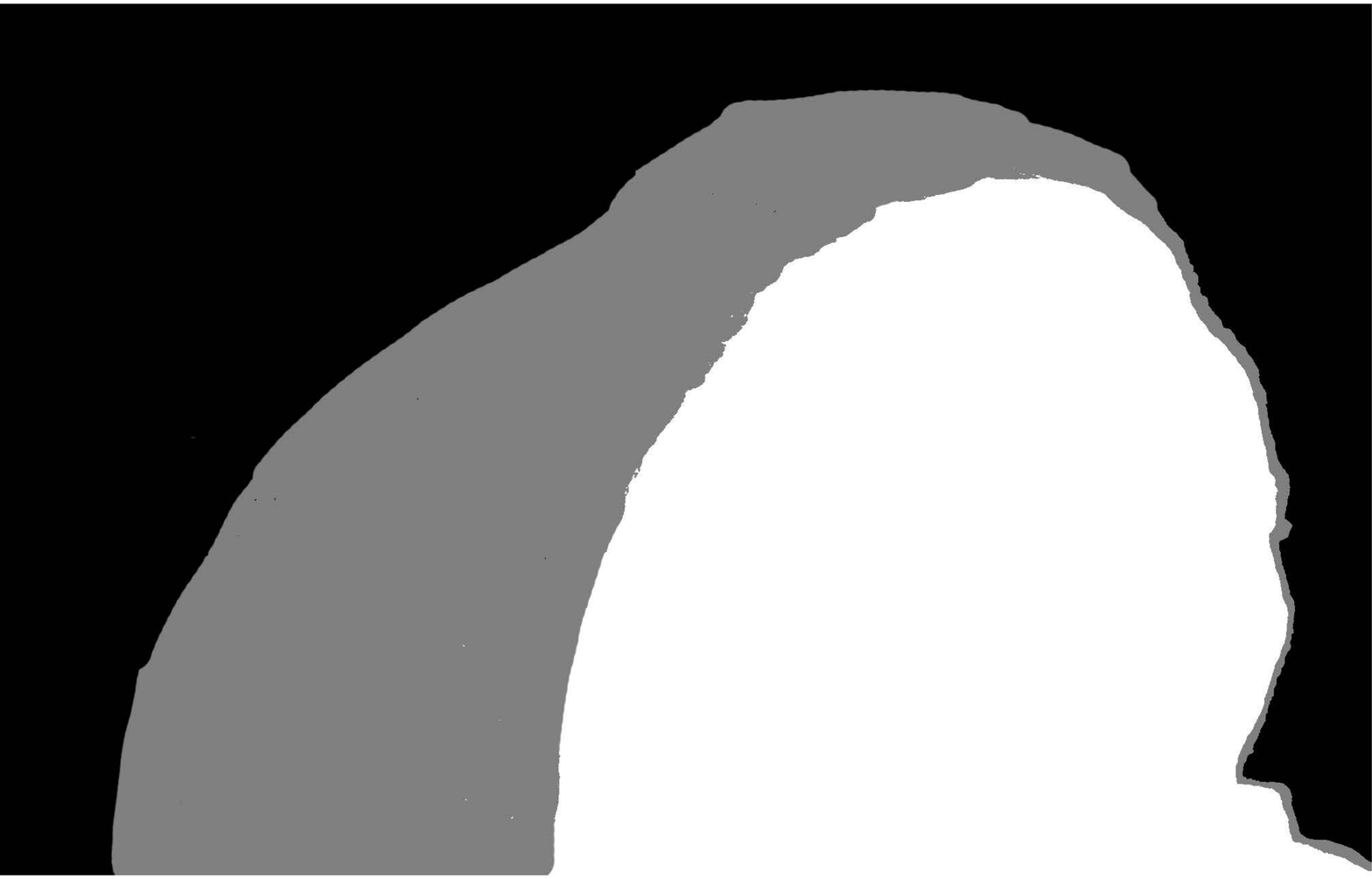}\vspace{-.08cm}
\includegraphics[width=1\linewidth]{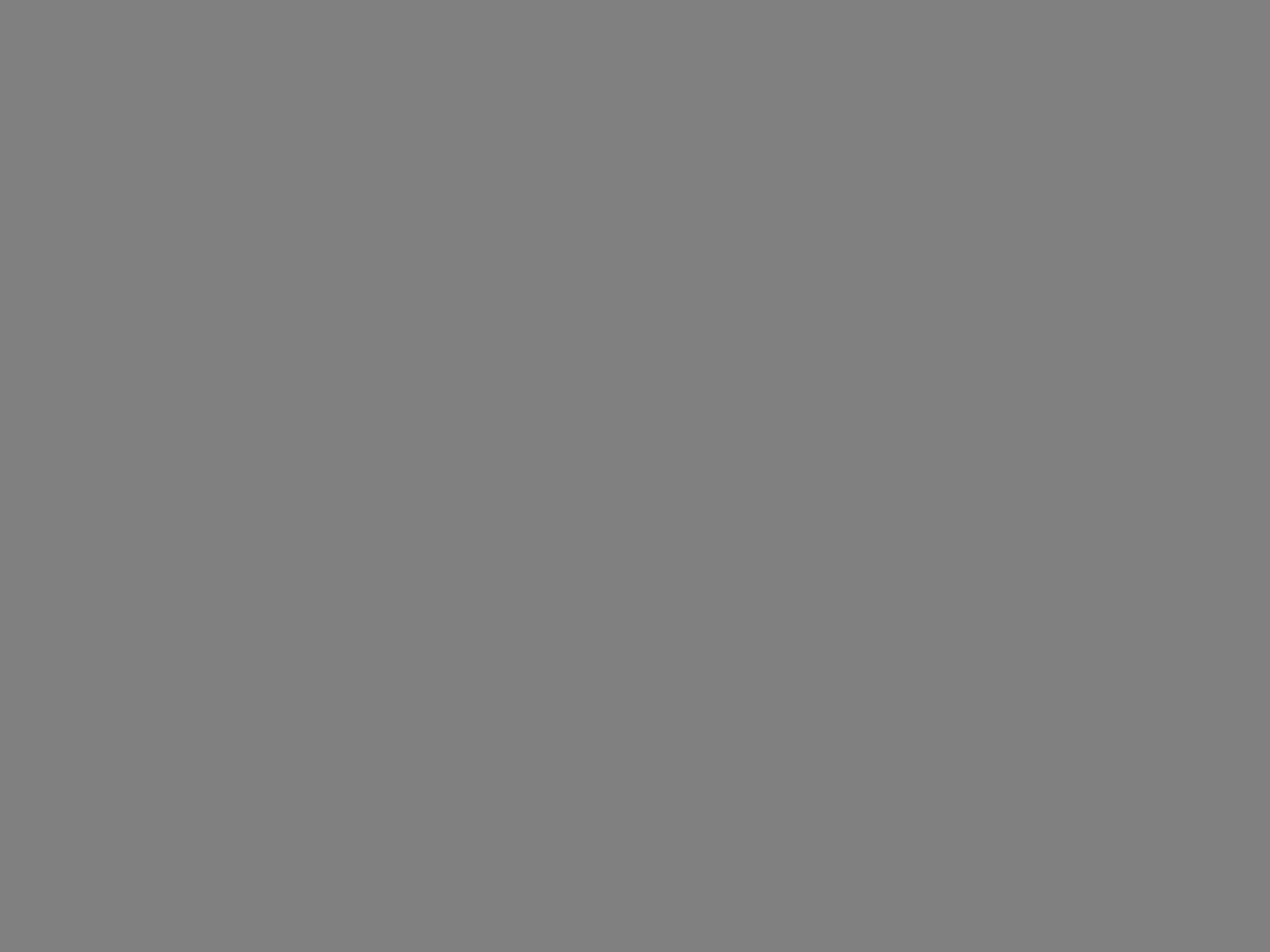}\vspace{-.25cm}
\label{img}   
 \end{minipage}}\hspace{-.15cm}
\subfigure[]{
\begin{minipage}[t]{0.245\linewidth} 
\includegraphics[width=1\linewidth]{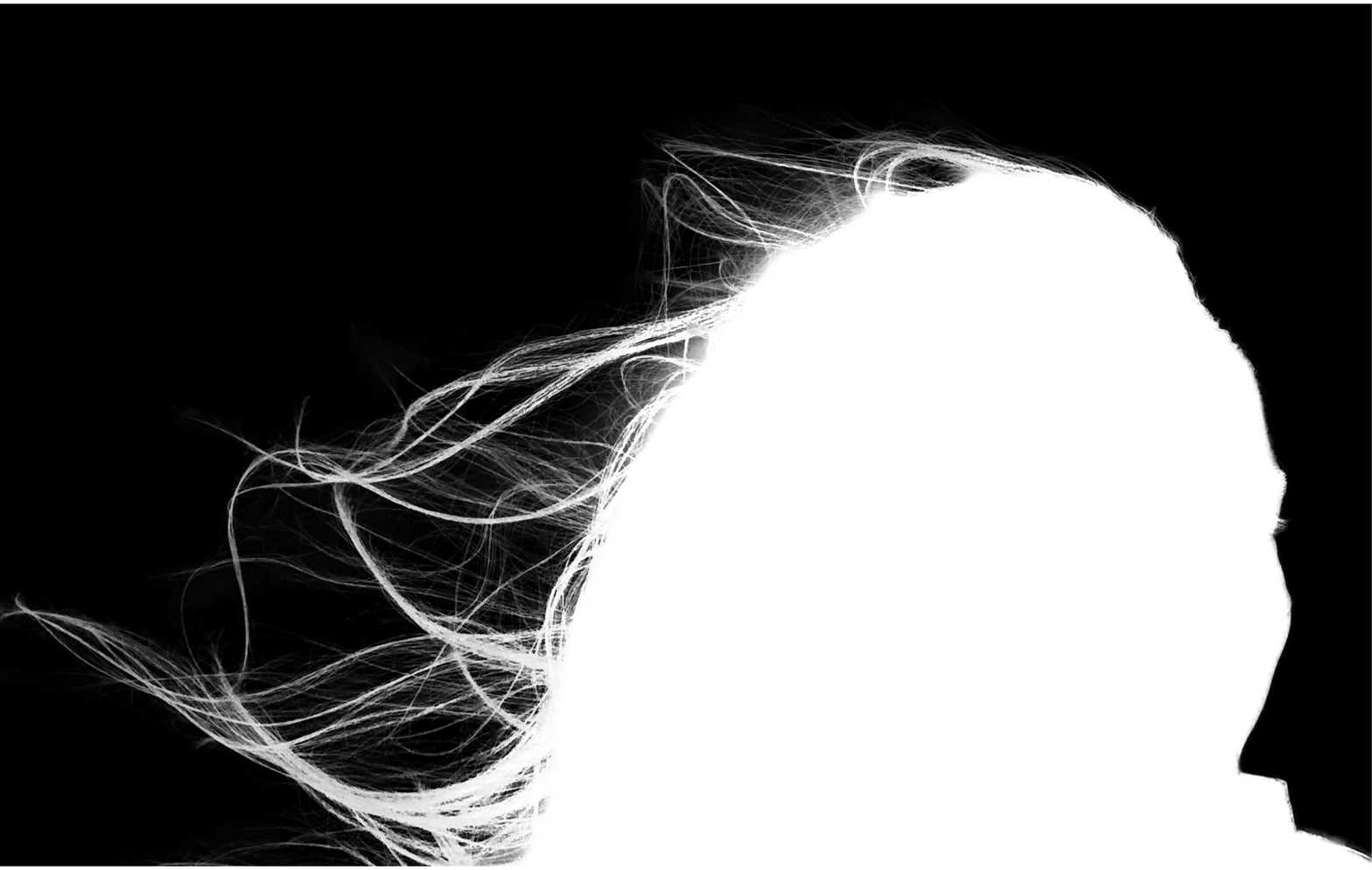}\vspace{-.08cm}
\includegraphics[width=1\linewidth]{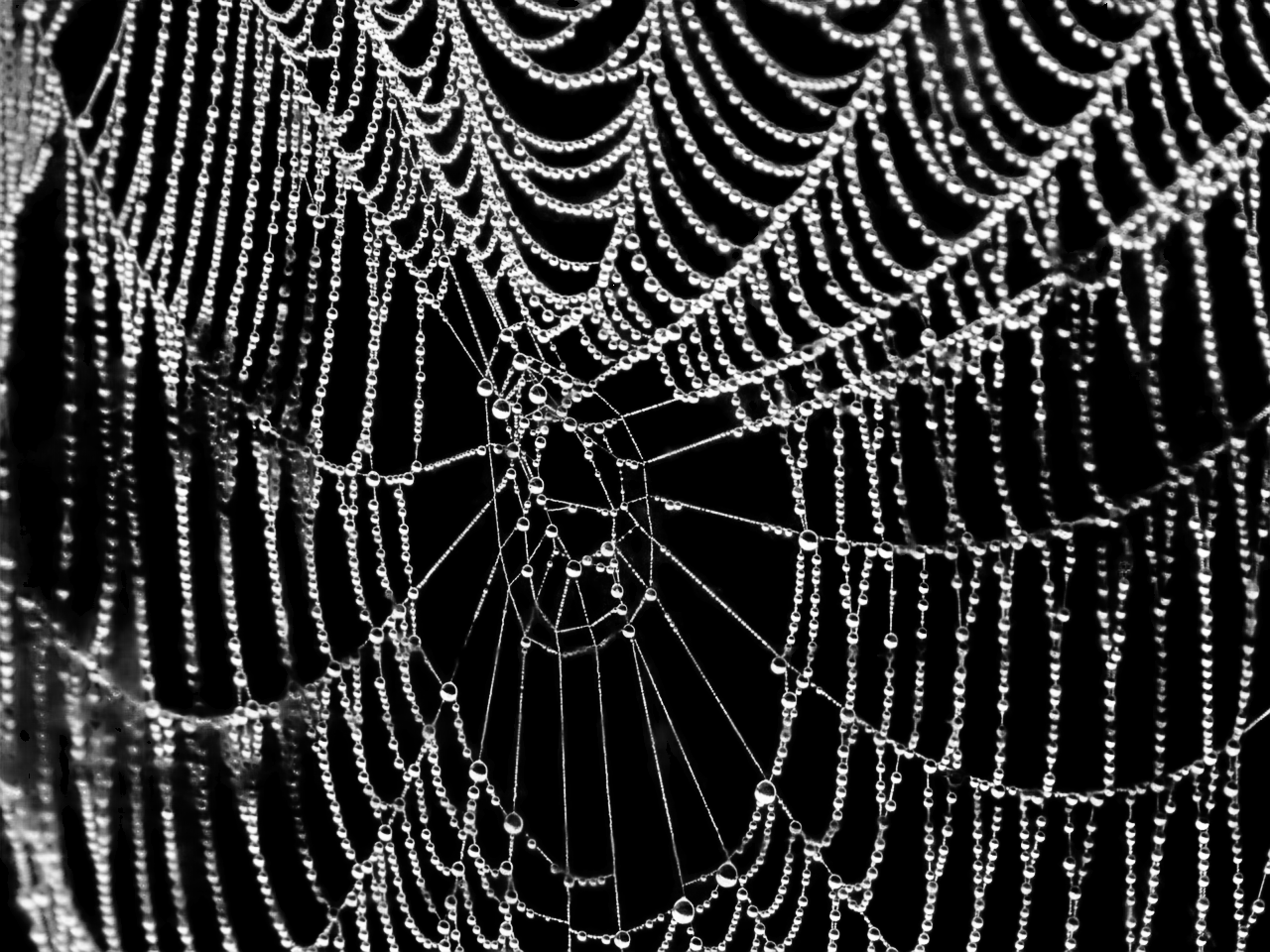}\vspace{-.25cm}
\label{img}   
\end{minipage}}\hspace{-.15cm}
\subfigure[]{
 \begin{minipage}[t]{0.245\linewidth}   
 \includegraphics[width=1\linewidth]{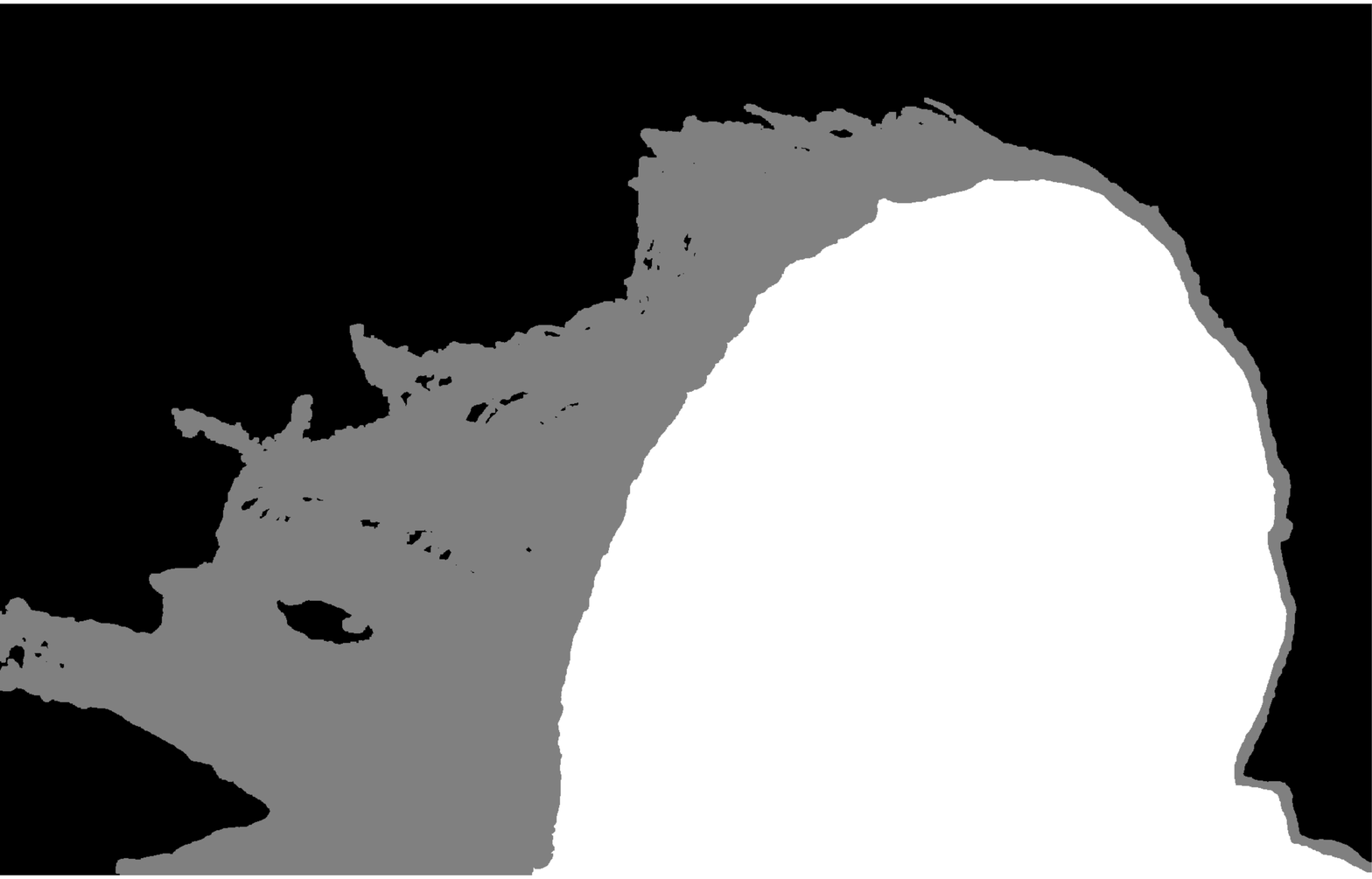}\vspace{-.08cm}
\includegraphics[width=1\linewidth]{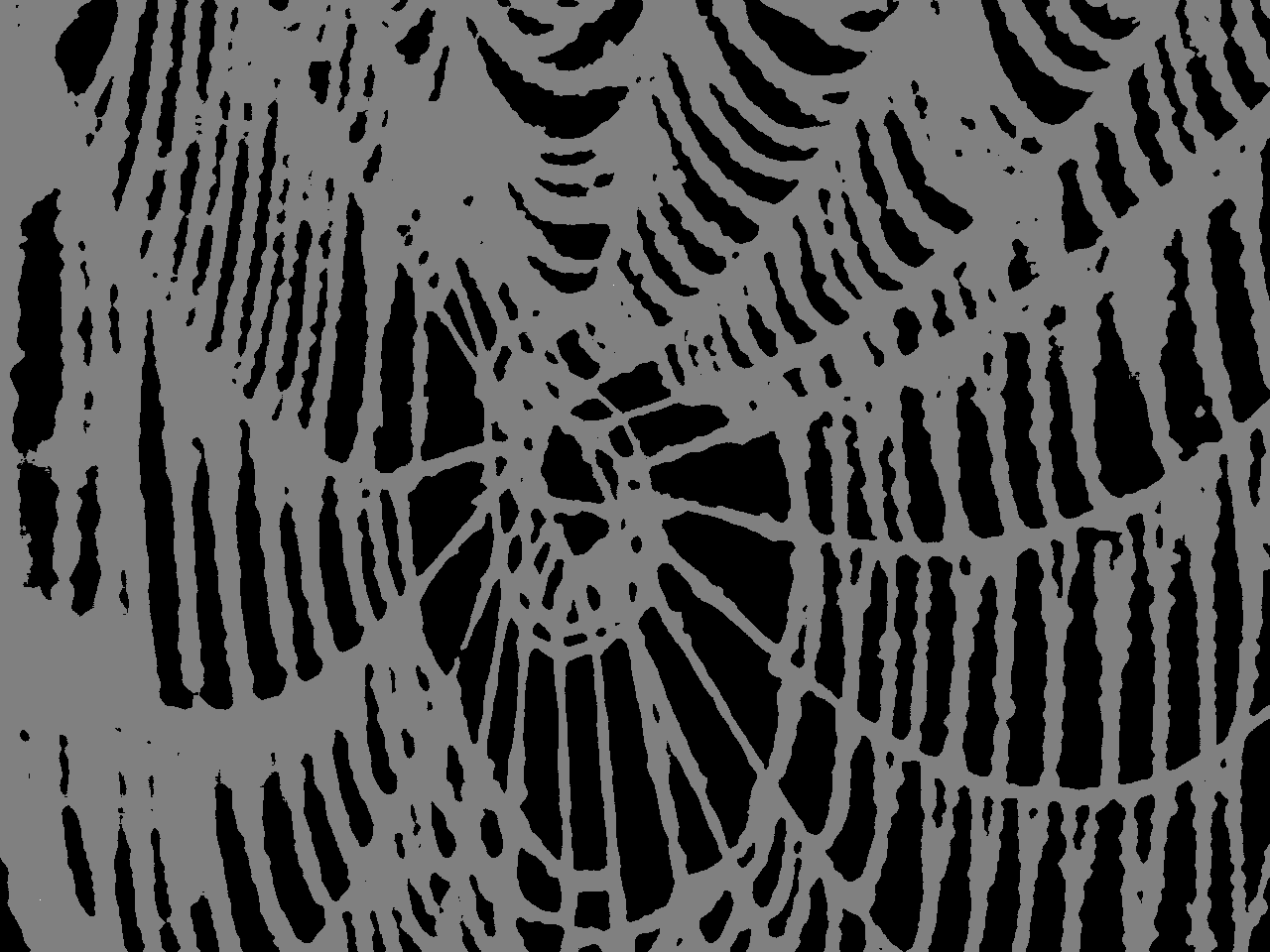}\vspace{-.25cm}
 \label{img}    
\end{minipage}}\hspace{-.15cm}
\caption{(a) Input image, (b) Input trimap, (c) Our matting results, (d) Corresponding trimap adaptation results. Row 1: the input trimap (b) from low-quality user labelling contains errors. Row 2: the input trimap (b) is a failure case of the Graphcut algorithm where all pixels are marked unknown. In both cases, the proposed method can produce reliable alpha mattes.}
\label{fig:teaser}
\vspace{-0.5cm}
\end{figure}

Image matting refers to the problem of accurately estimating the foreground object opacity in images and video sequences. It serves as a prerequisite for a broad set of applications, including film production and digital image editing. Formally, the input image $I$ is modeled as a linear combination of the foreground and background colors as follows~\cite{chuang2001bayesian}:

\vspace{-0.3cm}
\begin{equation}
I_{i} = \alpha_{i} F_{i} + (1-\alpha_{i} )B_{i}, \ \alpha_{i} \in [0, 1],
\label{eq:matting}
\end{equation}
\vspace{-0.3cm}

where $F_{i}$, $B_{i}$ and $\alpha_{i}$ denote the foreground, background color and alpha matte estimation at pixel $i$ respectively. Given an input image $I$, image matting aims to solve $F$, $B$, and $\alpha$ simultaneously. The problem is highly ill-posed, as according to Eq. \ref{eq:matting}, for an RGB image, 7 values are to be solved but only 3 values are known for each pixel. For most existing matting algorithms, the essential input that constrains the solution space is the trimap, a rough segmentation indicating the opaque and unknown regions. The trimap is generated either interactively by user scribbles, or automatically from binary image segmentation results. In either case, the input trimap is usually coarse, i.e., its unknown region (the gray region in Fig.~\ref{fig:teaser}b) contains both real semi-transparent pixels as well as a large amount of opaque ones. 
This is because providing an accurate trimap is tedious for manual labeling, and is difficult to achieve using existing image segmentation methods that run on low-resolution images. 

Unfortunately, previous image matting methods often ignore the inaccuracy of the input trimap, and try to directly estimate a good alpha matte from it. We argue that there is a classification problem that is not adequately addressed in this process. If we take a closer look at the trimap, pixels in the unknown region fall into three different sets: opaque foreground, opaque background, and the semi-transparent region. We call the first two types \textit{opaque pixels}, and the last type \textit{blended pixels}. The desired behavior of an image matting method is to produce exact 0s or 1s for the opaque pixels, while accurately estimating the fractional opacity (ranged between 0 and 1) for the blended pixels. From this perspective, two related but inherently different tasks are implied in image matting. The first is to classify pixels in the unknown region to iden blended pixel, and we name this task as \textit{trimap adaptation}. The second is to accurately calculate the opacity values for blended pixels, which we call \textit{alpha estimation}.



We observe that these two tasks demand quite different abilities from the algorithm. Trimap adaptation relies more on a good semantic understanding of the object shape and structure, so that it could effectively identify foreground and background regions in the unknown region based on image features. For alpha estimation, careful low-level exploitation of the photometric cues is more critical. Furthermore, trimap adaptation can be modeled as a classification task, and alpha estimation can be viewed as a typical regression task. Most of the existing image matting methods consider image matting as a single regression task, which ignores the classification nature resided in image matting. This observation brings us the question of how to reconcile the two very different aspects of the matting problem in one integrated solution.



Furthermore, existing matting methods, especially optimization-based ones, rely too much on low-level features such as color distributions and local textures, and lack the ability of incorporating high-level semantics. As shown in recent works \cite{xu2017deep,aksoy2018semantic}, inducing better understanding towards object shape and structure could help image matting. Although claiming to utilize high-level features, these methods typically rely solely on pretrained features and do not use explicit semantic objective as guidance. This is witnessed by the incomplete object structure extracted by the existing methods from areas where the background color is similar to the foreground object.

Motivated by the new observation that image matting should be disentangled into trimap adaptation and alpha estimation, we propose a simple yet powerful image matting framework named AdaMatting (\textbf{Ada}ptation and \textbf{Matting}), which resolves the limitations discussed above at the same time. AdaMatting performs trimap adaptation and alpha estimation within two distinct decoder branches in a multi-tasking manner. By explicitly supervising the model to distinguish blended pixels from opaque ones, and then using the refined trimap to afterwards constrain the alpha estimation output, the two branches separately handles the two different aspects of the task. Furthermore, the understanding towards object shape and structure information of the model is greatly enhanced by sharing features from the two tasks. See Fig.~\ref{fig:pipeline} for our detailed pipeline.

Our major contributions can be summarized as follows:
\vspace{-0.2cm}
\begin{itemize}
\item{We provide a new perspective that image matting should be disentangled into two tasks, namely \textit{trimap adaptation} and \textit{alpha estimation}, and demonstrate that the disentanglement of the two tasks is essential for improving performance of CNN-based image matting models.}
\vspace{-0.2cm}
\item{Following the new perspective, we propose a novel pipeline where trimap adaptation and alpha estimation are jointly optimized in a multi-tasking manner. Extensive experiments show that the proposed pipeline can better use semantic information to provide additional structural awareness and trimap fault-tolerance to the trained CNN model.}
\vspace{-0.2cm}
\item{The proposed method refreshes the state-of-the-art results on the most commonly used dataset Adobe Composition-1k~\cite{xu2017deep}, and ranks 1st on alphamatting.com~\cite{alphamatting}.}
\vspace{-0.2cm}

\end{itemize}

\section{Related Work}
\textbf{Natural Image Matting}
Natural image matting is essentially the per-pixel opacity estimation of the foreground region. The typical input to natural matting algorithms is in the form of \textit{scribbles} \cite{wang2005iterative} or \textit{trimaps} \cite{chuang2001bayesian}, which help reduce the solution space of this ill-posed problem.

Existing traditional methods can be categorized into color sampling based and alpha propagation based methods. Color sampling based methods \cite{chuang2001bayesian,gastal2010shared,he2011global,feng2016cluster} collect a set of known foreground and background samples to find candidate colors for a given pixel's foreground and background. According to the local smoothness assumption on the image statistics, these sampling colors are supposed to be ``close" to the true foreground and background colors. Once the foreground color and background color are determined, we can calculate the corresponding alpha value base on Eq. \ref{eq:matting}. Following this assumption, various sampling-based methods are proposed, including Bayesian matting \cite{chuang2001bayesian}, shared sampling matting \cite{gastal2010shared}, global sampling matting \cite{he2011global}, and sparse coding matting \cite{feng2016cluster}.

Compared to sampling-based methods, propagation-based approaches \cite{sun2004poisson,grady2005random,chen2013knn,aksoy2017designing,levin2008closed} avoid matte discontinuities which sampling-based approaches may suffer from. These methods utilize the affinities of neighboring pixels to propagate alpha values from the known regions into unknown ones. A popular approach among these is the closed-form matting \cite{levin2008closed}, which finds globally optimal alpha matte by solving a sparse linear system of equations. Other propagation-based approaches include Poisson matting \cite{sun2004poisson}, random walk matting \cite{grady2005random}, KNN matting \cite{chen2013knn} and information-flow matting \cite{aksoy2017designing}. 

Recently, deep learning has shown impressive performance on various computer vision tasks including image matting. Cho et al. \cite{cho2016natural} proposed an end-to-end architecture named DCNN that utilizes the results of closed-form matting \cite{levin2008closed} and KNN matting \cite{chen2013knn} to predict better alpha mattes. Shen et al. \cite{shen2016deep} proposed a fully automatic matting system for portrait photos based on end-to-end CNNs. Lutz et al. utilized \cite{lutz2018alphagan} the power of adversarial learning to extract alpha mattes which led to visually pleasing compositions. Wang et al. \cite{wang2018deep} showed a semantic-level pairwise similarity for propagation based matting can be learned via deep learning mechanism.

\textbf{Trimap Generation}
To the best of our knowledge, there are no existing work for trimap adaptation (\textit{i.e.} generating the accurate optimal trimaps). The most related topic is automatic trimap generation. Automatic trimap generation has been an important part for traditional matting methods. Wang et al. \cite{wang2007automatic} used depth information acquired by a time-of-flight range scanner to obtain trimap. Some other algorithms \cite{cho2017automatic,hsieh2013automatic} rely on the binary segmentation to obtain the coarse trimaps. \cite{singh2013automatic} used the RGB image feature maps together with morphological dilation to automatically generate trimap, and refined trimap using region growing mechanism. \cite{al2015novel} first introduced the Gestalt laws to the matting problem, making more robust trimap generation possible. More recently, \cite{chen2018semantic, shen2016deep} utilized neural networks to generate trimaps, greatly improving the matting performances.

\textbf{Multi-task Learning}
Multi-task learning is a sub-field of machine learning, in which multiple learning tasks are solved within a single model simultaneously. Compared with training separate models for each task, multi-task learning improves learning efficiency and prediction accuracy for each task by utilizing their inter-relation. In computer vision, there are various exemplars of using multi-task learning, \textit{e.g.} joint object detection and semantic segmentation \cite{hariharan2014simultaneous}, simultaneous depth estimation and scene parsing \cite{xu2018pad}, and universal network for handle low, middle and high-level vision tasks \cite{kokkinos2017ubernet}. Recently, Kendall et al. \cite{kendall2017multi} proposed a general way of combining multiple loss functions to simultaneously learn multiple objectives using homoscedastic task uncertainty. By dynamically adjusting the weights for each objective, their model could obtain superior performance compared to separately trained models.

\vspace{-0.2cm}
\section{Method}
\vspace{-0.2cm}
According to the disentangled view of image matting aforementioned, two related but subtly different tasks are implied in image matting, namely the trimap adaptation, a classification task, and the alpha estimation, a regression task. We propose a novel pipeline for image matting, in which the two sub-tasks are solved simultaneously in a multi-task learning manner, and then the final mattes are propagated based on the results of the sub-tasks.

In this section, we first formulate the task of trimap adaptation, and then describe the pipeline and training schemes of our proposed AdaMatting (\textbf{Ada}ptation and \textbf{Matting}).

\begin{figure*}[!htp]
\centering
\centerline{\includegraphics[width=17.5cm]{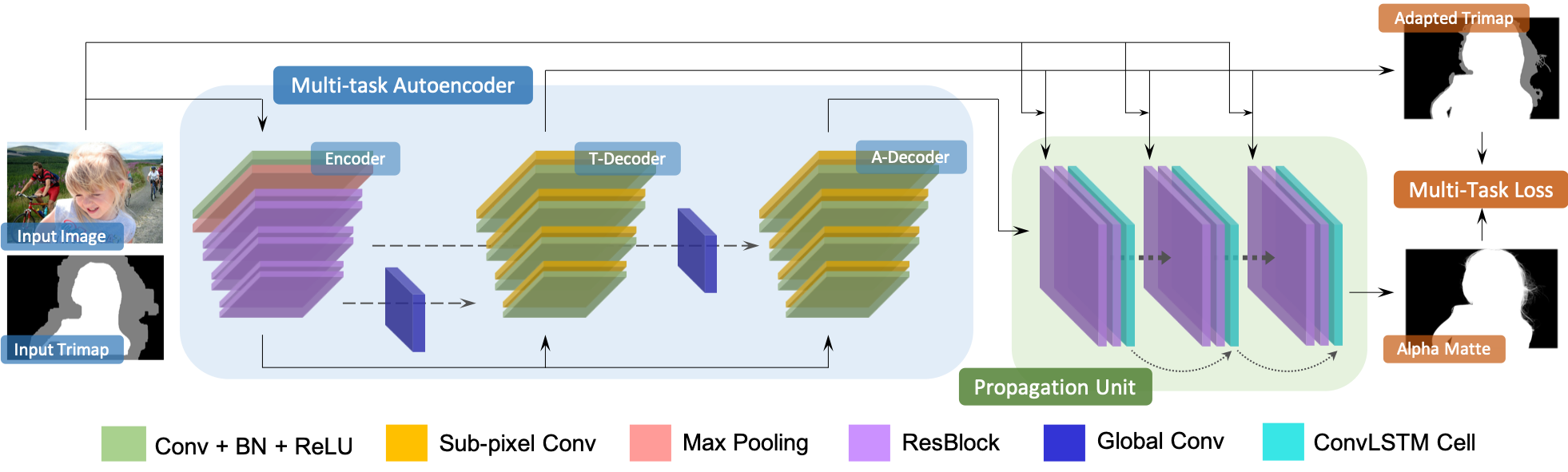}}
\caption{Pipeline of the proposed AdaMatting. T-Decoder and A-Decoder stand for trimap decoder and alpha decoder, respectively. Symmetric shortcuts are linked to different levels of layer for the two decoders.}
\label{fig:pipeline}
\vspace{-0.5cm}
\end{figure*}

\subsection{Trimap Adaptation}
\label{sec:trimap_adaptation}


We begin by formally defining the task of trimap adaptation. Let $\alpha_{gt}$ be the ground truth alpha mattes. The corresponding \textit{optimal trimap} $ T_{opt}$ of an image can be naturally defined as:

\vspace{-0.7cm}
\begin{equation}
 T_{opt}(x, y)=
\begin{cases}
background  & \text{if $\alpha_{gt}(x, y) = 0$,} \\
unknown  & \text{if $0 < \alpha_{gt}(x, y) < 1$,} \\
foreground  &\text{if $\alpha_{gt}(x, y) = 1$,} \\
\end{cases}
\label{eq:opt_trimap}
\end{equation}
where $(x, y)$ stands for each pixel location on the image. Given an input image conditioned with a trimap (which could be coarse), the trimap adaptation aims to predict the optimal trimap $T_{opt}$. Intuitively, in trimap adaptation, we separate the semitransparent regions from the opaque foreground and background. This is reminiscent of the semantic segmentation task which also divides the image into discrete parts. From the $T_{opt}$ defined above, the image matting task naturally factors into two steps: (1) deciding if the alpha should be exactly zero, one or neither, (2) computing the exact alpha if the region is considered semitransparent. Note that we do not require the predicted label to be strictly compatible with the input trimap: if the user input contains minor error, we would like our model to correct it.


There are several reasons why separating trimap adaptation and alpha estimation is helpful. First, the two tasks require different training strategies and mode-of-operation of the model. Also, the classification task and the regression task usually desire different loss objectives. Therefore, on one hand, separating the two tasks relieves the burden of the regressor to generate exact zero or one values for the opaque pixels to a great extent. On the other hand, when the exact fractional value of $\alpha$ is hidden, the semantic and structural information of the object is expected to be more eminent for the classifier to exploit. Second, our final results are propagated using the predicted $\tilde{T}$ instead of the coarse input trimap, making our model more robust and fault-tolerant to the coarse input trimap.



Fig. \ref{fig:teaser} (d) shows examples of trimap adaptation performed by our model. As can be observed, the unknown region in the first input trimap is wide and erroneous, not covering all of the hair due to low-quality labeling. After performing trimap adaptation, the output trimap is not only narrowed but also corrected, resulting in more credible alpha mattes. For the second input image (``cobweb'' of the Adobe Composition-1k testing set), the automatic trimap generation algorithm (Graphcut \cite{greig1989exact} based) fails to provide a meaningful trimap. However, the proposed AdaMatting can surprisingly adapt a rather precise trimap under this extreme condition, thus perfectly solves this hard case of image matting.

\subsection{Network Overview}\vspace{-0.2cm}
As mentioned above, the trimap adaptation requires more semantic understanding of the object shape and structure, and the image matting relies more on careful low-level exploitation of the photometric cues. Solving these two tasks simultaneously, while sharing intermediate representations, can reasonably enhance the performance of the entire model. Hence we designed a fully end-to-end CNN model named AdaMatting. Fig. \ref{fig:pipeline} depicts the pipeline of AdaMatting, which consists of one encoder producing shared representations, followed by two dependent decoders, solving trimap adaptation and alpha estimation respectively. The result of trimap adaptation and the intermediate alpha matte are then sent into the propagation unit, forming the final alpha matte.

The proposed AdaMatting takes an image concatenated with the corresponding trimap as input. First, a front-end fully convolutional encoder (adopted from the ResNet-50~\cite{he2016deep}) produces deep features as shared representations. Then two separate decoders are employed for each task, aiming to learn mappings from the shared representations to the desired output. Each decoder consists of several 3 $\times$ 3 convolutional layers and up-sampling modules. The trimap decoder outputs 3-channel classification logits, guided by the cross-entropy loss. The alpha decoder outputs a 1-channel intermediate alpha estimation, which is forwarded into the propagation unit for further refinement.

Detailed network architecture is depicted in Fig. \ref{fig:pipeline}. Here we explain the major modules of our model:


\textbf{Multi-task Autoencoder}
The primary module of our pipeline is the multi-task autoencoder, designed based on the widely used U-Net architecture, as it has achieved a great success for numerous computer vision tasks~\cite{ronneberger2015u,zhang2018dynamically,zhang2018dmcnn}. According to the observation that the trimap adaptation relies more on high-level features and the the alpha estimation relies more on low-level ones, the symmetric shortcuts are linked between different levels of layer for the two decoder. More specifically, the trimap decoder employs deep and middle layer symmetric shortcuts, and the alpha decoder employs middle and shallow layer symmetric shortcuts. Also, recent works~\cite{peng2017large,zhou2014object} show that the size of effective receptive field plays an important role on segmentation tasks. To further enlarge the receptive field while keeping acceptable computational costs, global convolutions~\cite{peng2017large} are employed on the shortcuts. This modification further enlarges the receptive field, contributing to more reliable and locally consistent results.

\begin{figure}
\centering
\subfigure{
 \begin{minipage}[t]{0.31\linewidth}   
 \includegraphics[width=1\linewidth]{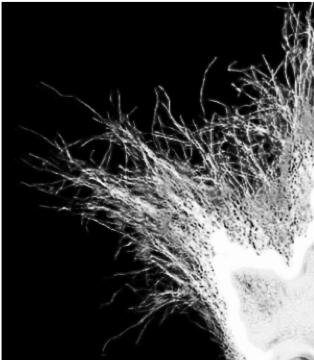}
 \centerline{\small Step 1}\vspace{-0.2cm}
 \end{minipage}}
 \subfigure{
 \begin{minipage}[t]{0.31\linewidth}   
 \includegraphics[width=1\linewidth]{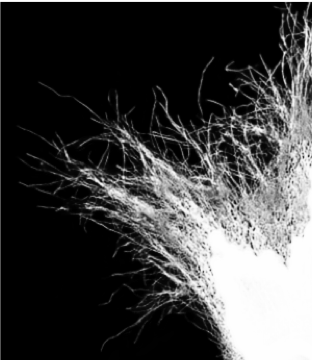}
 \centerline{\small Step 2}\vspace{-0.2cm}
 \end{minipage}}
 \subfigure{
 \begin{minipage}[t]{0.31\linewidth}   
 \includegraphics[width=1\linewidth]{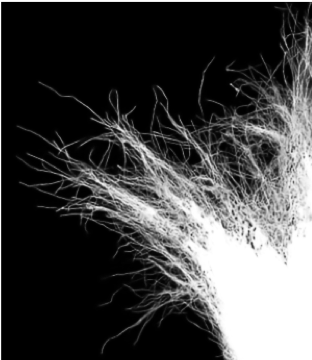}
 \centerline{\small Step 3}
 \end{minipage}}
\caption{Visualization of the propagated intermediate alpha mattes. Artifacts are gradually removed and the hair progressively becomes clearer and more distinguishable.}
\label{fig:propagation}
\vspace{-0.7cm}
\end{figure}

\textbf{Propagation Unit}
Inspired by the widely used propagation-based approaches~\cite{sun2004poisson,grady2005random,chen2013knn,aksoy2017designing,levin2008closed}, we designed a propagation unit empowered by recent advance of the convolutional long short term memory (LSTM) networks~\cite{xingjian2015convolutional}. As shown in our pipeline in Fig.~\ref{fig:pipeline}, the unit is composed of two ResBlocks~\cite{he2016deep} and a convolutional LSTM cell. In each recurrent iteration, the input image, the adapted trimap, and the previous alpha matte propagation result are taken as input. The ResBlocks extract features from the inputs, while the convolutional LSTM cell keeps memory between the propagation steps.

Similar to the traditional propagation based methods, the propagation unit progressively refines predicted alpha mattes, yielding final results with more accurate edge details and significantly less undesired artifacts. Fig.~\ref{fig:propagation} illustrates an examples of how alpha matte is refined within the designed propagation unit. As can be seen, the hair progressively becomes more clearly distinguishable. Also, the blurring artifacts are eliminated within the propagation process.



\vspace{-0.2cm}
\subsection{Multi-task Loss}
Multi-task learning aims to solve multiple tasks within one model, while achieving superior efficiency and performance compared to separately trained models. It can be considered as an approach to induce knowledge transfer by sharing the domain information between complementary tasks \cite{uhrig2016pixel, kendall2017multi}. From an implementation aspect, by utilizing shared representations and designed objectives, multiple tasks are capable of learning from each other in an effective and efficient way.

Specifically in the AdaMatting, the two tasks are trimap adaptation and alpha estimation. Trimap adaptation, as mentioned above, can be modelled as a segmentation task, splitting the input images into solid foreground, solid background, and semitransparent regions. The process of solving such kind of segmentation problem could lead to rich semantic features, which help solve the alpha matte regression in return. 

Instead of a linearly combined loss, we adopt the task uncertainty loss~\cite{kendall2017multi}. Our loss can be formulated as:

\vspace{-0.6cm}
\begin{equation}
\begin{split}
\mathcal{L}(\{\tilde{T}, \tilde{\alpha}\}, \{T_{opt}, \alpha_{gt}\}) &= \frac{1}{2\sigma_1^2} \mathcal{L}_T(\tilde{T}, T_{opt}) \\
&+ \frac{1}{\sigma_2} \mathcal{L}_\alpha(\{\tilde{T}, \tilde{\alpha}\}, \alpha_{gt}) + \log{2\sigma_1\sigma_2}, \vspace{-0.6cm} \\
\end{split}
\label{eq:mtloss}
\end{equation}
where $\tilde{T}$ and $\tilde{\alpha}$ stand for the output of trimap adaptation and alpha estimation, $\sigma_1$ and $\sigma_2$ stand for dynamically adjusted task weights, $\mathcal{L}_T$ and $\mathcal{L}_\alpha$ stand for trimap adaptation loss and alpha estimation loss, respectively. More specifically, $\mathcal{L}_T$ is the cross-entropy loss, and $\mathcal{L}_\alpha$ is the $L_1$ loss, calculated on the unknown regions of $\tilde{T}$ (denote as $\tilde{T}_u$) only:

\vspace{-0.5cm}
\begin{equation}
\mathcal{L}_{\alpha}(\{\tilde{T}, \tilde{\alpha}\}, \alpha_{gt}) =  \frac{1}{|\tilde{T}_u|} \sum_{s \in \tilde{T}_u}{|\tilde{\alpha}(s) - \alpha_{gt}(s)|},
\vspace{-0.3cm}
\end{equation}
where $|\tilde{T}_u|$ is the number of pixels in $\tilde{T}_u$. The loss actually disentangles the image matting into two parts, as described in Section \ref{sec:trimap_adaptation}, assuring each decoder to learn structural semantics and photometric information respectively.

Note that the trade-off parameter of the two tasks are dynamically adjusted during training time by the back-propagation algorithm, which avoids the expensive and cumbersome searching process for the optimal weights.







\section{Experiments}

\begin{table*}
\begin{center}
\setlength{\tabcolsep}{1.2mm}
\caption{Average ranking results of our methods and 5 representative state-of-the-art techniques on the alphamatting.com dataset \cite{alphamatting}. Best results are shown in bold. S, L, U stand for different type of input trimaps. See \url{alphamatting.com} for details.}
\label{table:alphamatting}
\vspace{0.3cm}
\begin{tabular}{ccccc|cccc|cccc}
Methods &\multicolumn{4}{c}{Gradient Error} &\multicolumn{4}{c}{MSE} &\multicolumn{4}{c}{SAD}\\
\noalign{\smallskip}
\hline
\rowcolor{mygray}
& Overall & S & L & U  &Overall & S & L & U & Overall & S & L & U\\
AdaMatting (ours) &  \textbf{5.2}& \textbf{2.8}& \textbf{2.8}& \textbf{10}&           \textbf{5.3}& \textbf{3.8}& \textbf{4.8}& \textbf{7.5}          & \textbf{4.6}& \textbf{3.9}& \textbf{3.8}& \textbf{6.1}\\
\rowcolor{mygray}
SampleNet Matting  & 6.2 &3.1 &3.3 &12.1        &6.4	&4&6.4&8.9          &5.3	&3.9	&4.5	&7.4\\
AlphaGAN~\cite{Lutz2018} & 13.2 &12 &10.8 &16.8        &14.3&14.8&15.1&13.1       &11.2&12&11&10.6\\
\rowcolor{mygray}
DCNN~\cite{cho2016natural}  &14.6  & 17.9& 14.4& 11.6                   &10&11.6&7.9&10.5                     &10.5&12.5&8.6&10.4      \\
DIM~\cite{xu2017deep}  &14.3& 10.8& 11 &21                      &9.3&8	&8&11.9                          &7.1&8.3&6.1&6.9\\
\rowcolor{mygray}
IF~\cite{aksoy2017designing}  &16.4&19.5&14.1&15.1        &10&12.5	&9&8.6          &8.8&9.9	&8.9	&7.5 \\ 
\hline
\vspace{-1.1cm}
\end{tabular}
\end{center}
\end{table*}
\setlength{\tabcolsep}{1.4pt}

\begin{figure*}[!htp]
\begin{center}
\subfigure{
\begin{minipage}[t]{0.16\linewidth} 
\includegraphics[width=1\linewidth]{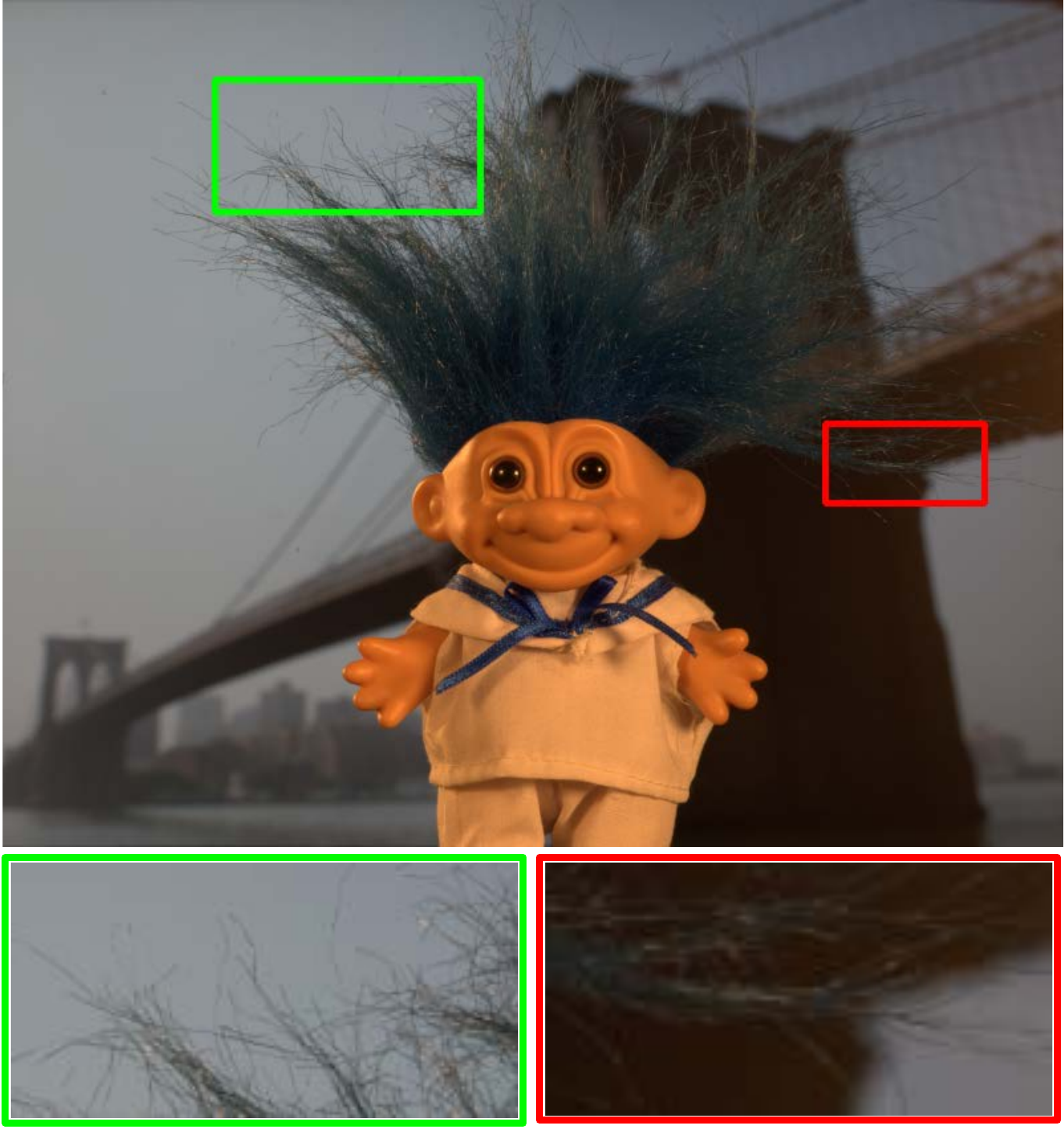}
\centerline{\small Troll}
\label{img}
\end{minipage}}\hspace{-.15cm}\vspace{-.1cm}
\subfigure{
\begin{minipage}[t]{0.16\linewidth}   
 \includegraphics[width=1\linewidth]{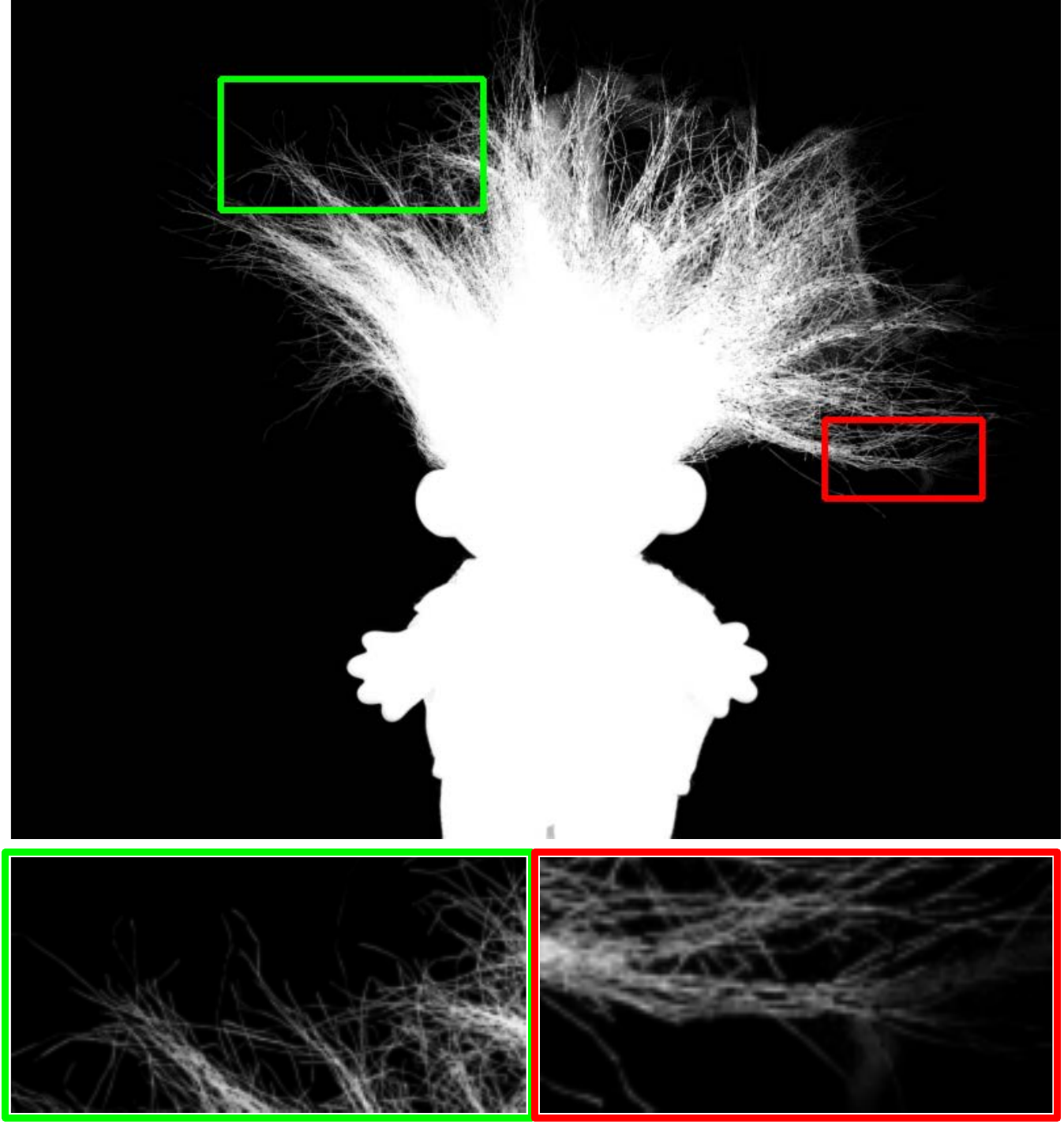} 
 \centerline{\small TLGM \cite{li2017three}}
 \label{img}    
 \end{minipage}}\hspace{-.15cm}\vspace{-.1cm}
  \subfigure{
\begin{minipage}[t]{0.16\linewidth} 
 \includegraphics[width=1\linewidth]{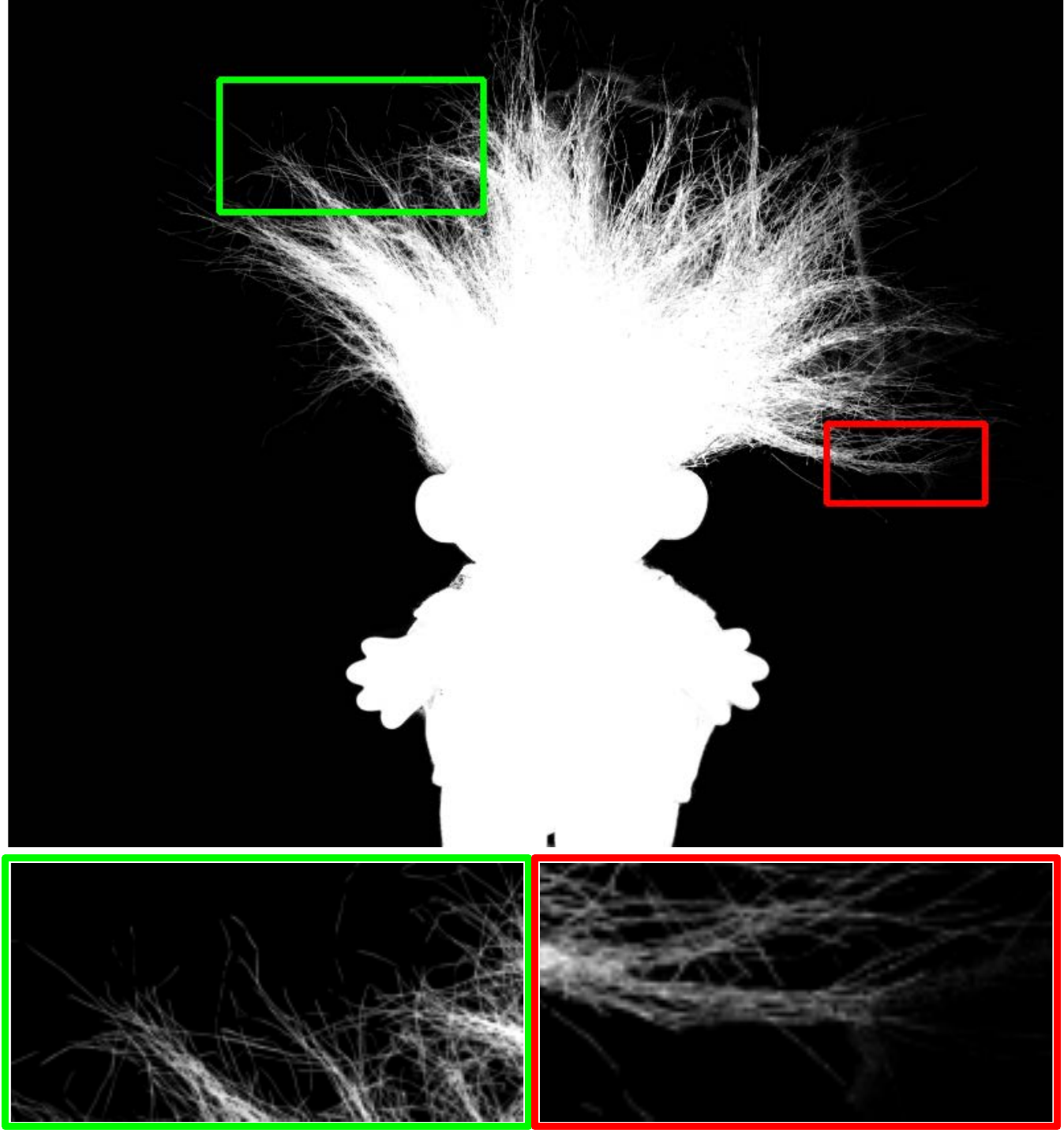} 
 \centerline{\small IF \cite{aksoy2017designing}}
 \label{img}    
 \end{minipage}}\hspace{-.15cm}\vspace{-.1cm}
 \subfigure{
\begin{minipage}[t]{0.16\linewidth}   
 \includegraphics[width=1\linewidth]{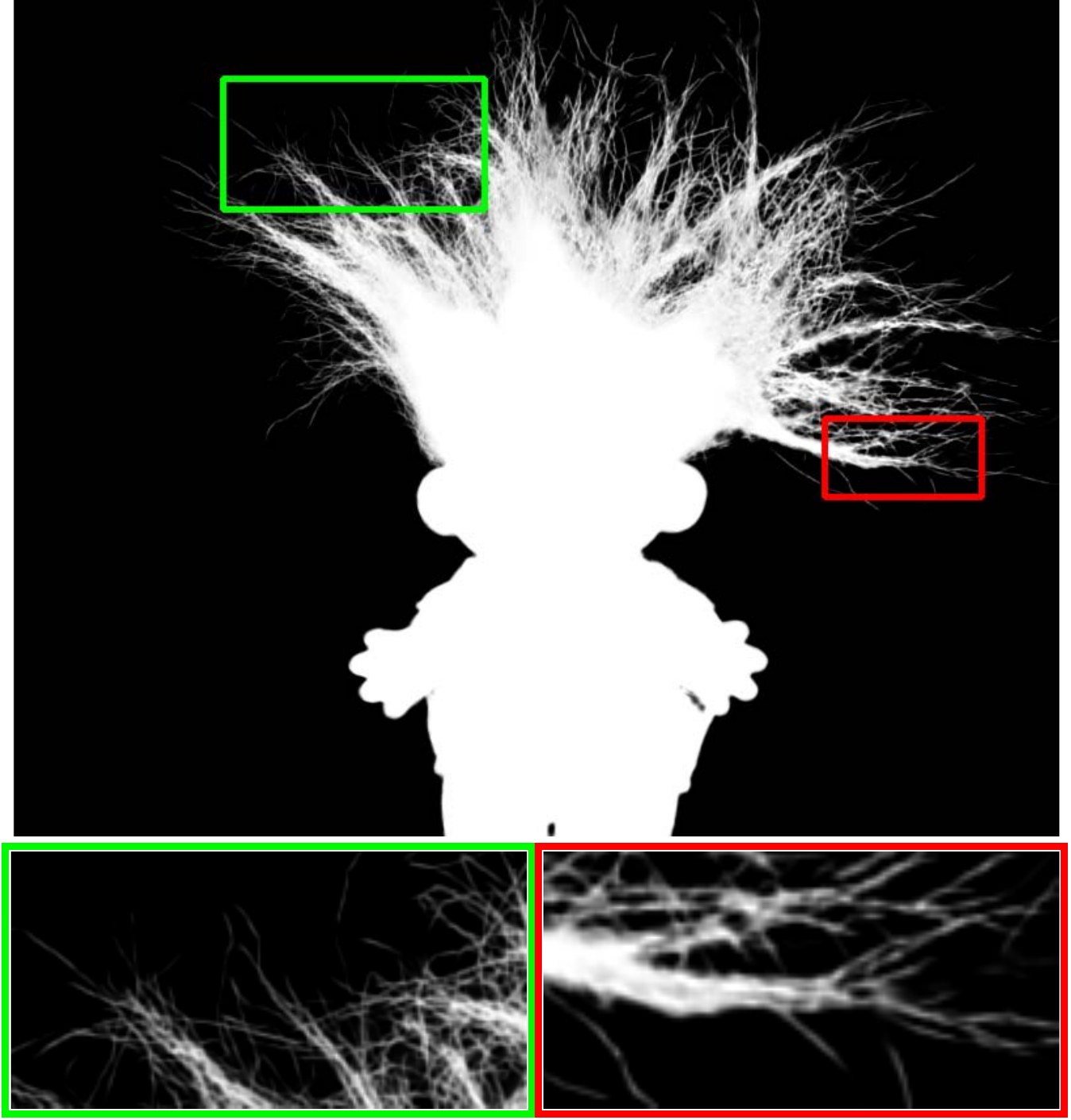} 
 \centerline{\small DIM \cite{xu2017deep}}
 \label{img}    
 \end{minipage}}\hspace{-.15cm}\vspace{-.1cm}
  \subfigure{
\begin{minipage}[t]{0.16\linewidth}   
 \includegraphics[width=1\linewidth]{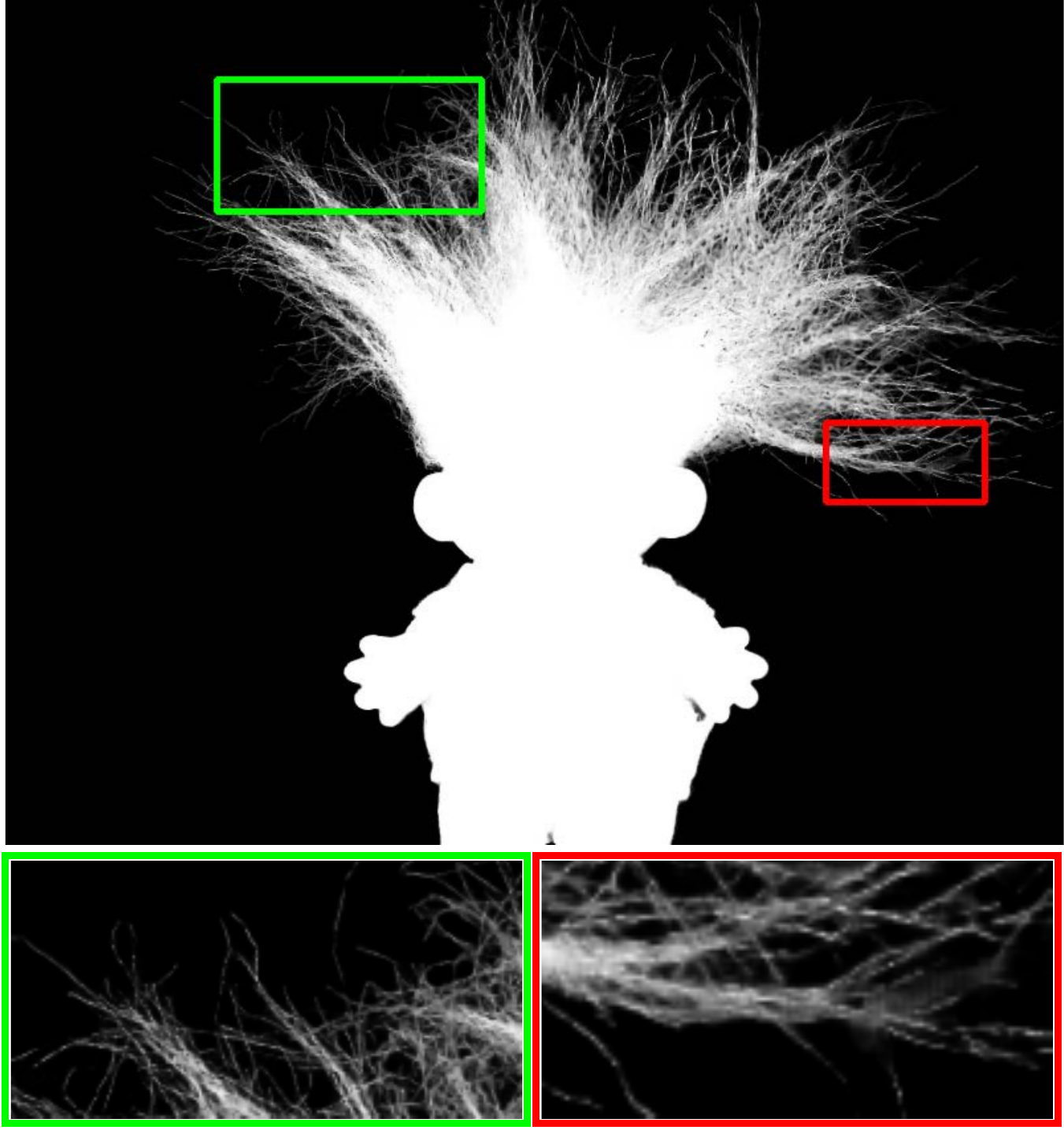} 
 \centerline{\small AlphaGAN \cite{lutz2018alphagan}}
 \label{img}    
 \end{minipage}}\hspace{-.15cm}\vspace{-.1cm}
 \subfigure{
\begin{minipage}[t]{0.16\linewidth}   
 \includegraphics[width=1\linewidth]{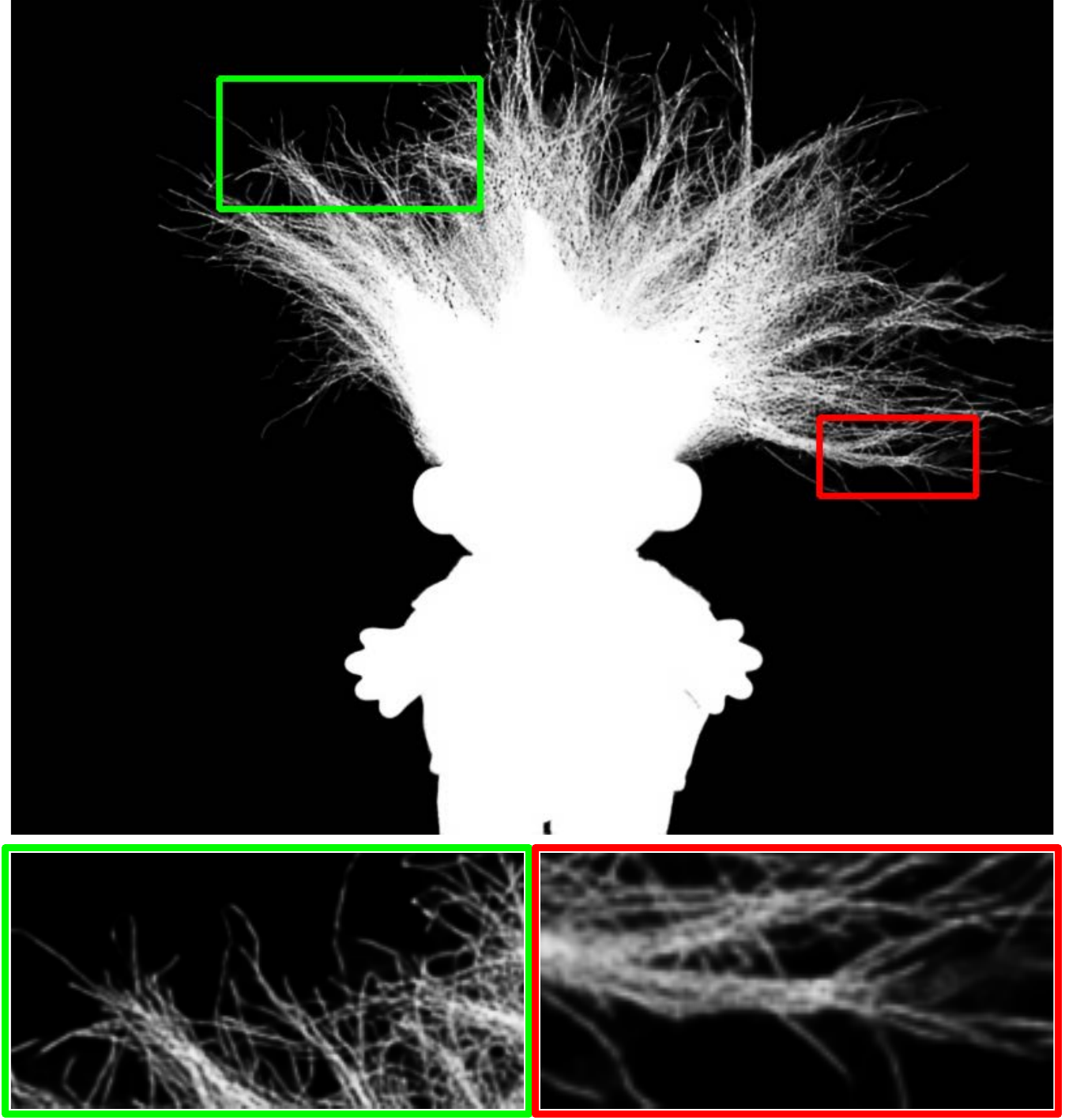} 
 \centerline{\small Ours}
 \label{img}    
 \end{minipage}}
  \subfigure{
 \begin{minipage}[t]{0.16\linewidth} 
 \label{img}    
 \end{minipage}}\end{center}
\caption{Qualitative comparisons on two images of alphamatting.com test set. \cite{alphamatting} The figure shows the alpha predictions of the test image ``troll" with trimap ``user".}
\label{fig:alphamatting}
 \vspace{-0.5cm}
\end{figure*}

\label{sec:blind}
We conduct extensive experiments and demonstrate the efficacy of our methods on two public datasets: (1) \textit{alphamatting.com} \cite{alphamatting} and (2) Adobe Composition-1k \cite{xu2017deep} test set. The latter one has a wider range of object types and more complicated background scenes. In this section, we compare our complete AdaMatting with current state-of-the-art methods both quantitatively and qualitatively.

\subsection{Experiment Settings}
\textbf{Datasets} The first dataset, \textit{alphamatting.com} \cite{alphamatting}, is a well-known online evaluation benchmark for natural image matting methods. It includes 27 training images and 8 testing images with 3 different kind of trimaps, namely,``small'', ``large'' and ``user'', representing different input trimap qualities. The second dataset is the \textit{Composition-1k} \cite{xu2017deep}, which provided 431 foreground images as well as their ground truth alpha mattes. 100 background images from COCO \cite{Andriluka2014} are selected for each foreground. We completely followed the composition order used by \cite{xu2017deep} while using the dataset.

\textbf{Evaluation Metrics} We use four quantitative metrics for matting evaluation. Namely the sum of absolute differences (SAD), mean square error (MSE) and the gradient error (Grad). Empirical studies show that Grad is better suited for perceptual comparisons of matting methods \cite{alphamatting}. 



\textbf{Implementation Details}
Inspired by \cite{chen2017rethinking}, we use the ``poly" learning rate policy where current learning rate is defined as the base learning rate multiplied by $(1- \frac{iter}{max\_iter})^{p}$. The base learning rate and $p$ is set to 0.0001 and 0.9 respectively. The Adam optimizer is used, with momentum and weight decay set to 0.9 and 0.0001 respectively. The $\sigma_1$ and $\sigma_2$ are both initialized to 4 in the multi-task loss.

For all experiments, we train for 120 epochs with a batch-size of 16. The training patches sized $800 \times 800$ to $320 \times 320$ along the unknown regions in the trimap are randomly cropped and then resized to $320 \times 320$ patches, as training with larger patches could introduce more semantic information. The training trimaps are generated from the ground truth alpha mattes using the random erode and dilate technique \cite{xu2017deep}. For data augmentation, we adopt random flip and random resize between 0.75 and 1.5 for all images, and additionally add random rotation between -45 and 45 degrees. Due to the difference between classification and image matting task, the model weights are not initialized by ImageNet pre-training as in \cite{xu2017deep,lutz2018alphagan}. The training data are randomly shuffled in each epoch. The training process takes about 2 days with eight NVIDIA TITAN X GPUs for each experiment. During inference, the full-resolution input images and corresponding trimaps are concatenated as the 4-channel input and fed into the network. The propagation unit recurs for 3 times.


\subsection{Results on alphamatting.com}

We submitted our result to the alphamatting.com \cite{alphamatting}. The AdaMatting achieves state-of-the-art performance, ranking the first for the average performance on all three metrics. The gradient error and the MSE results are shown in Tab. \ref{table:alphamatting}.

Several visual comparisons are shown in Fig. \ref{fig:alphamatting}. As can be seen from the figure, our results contain much more details compared to other state-of-the-arts. Specifically, for the ``Troll'' test image in the first row, we produce sharper details with less artifacts compared to other models.

\subsection{Results on Composition-1k}

For the Composition-1k test set, we evaluate 6 recent state-of-the-art methods, namely Closed Form \cite{levin2008closed}, KNN \cite{chen2013knn}, DCNN \cite{cho2016natural}, Information Flow \cite{aksoy2017designing}, AlphaGAN \cite{Lutz2018}, and Deep Image Matting \cite{xu2017deep}. The quantitative results under the Grad, SAD and MSE are shown in Table \ref{table:adobe}. Obviously, our model outperforms all other methods on all metrics by a large margin.

\begin{figure*}
\centering
\subfigure{
\begin{minipage}[t]{0.19\linewidth} 
\includegraphics[width=1\linewidth]{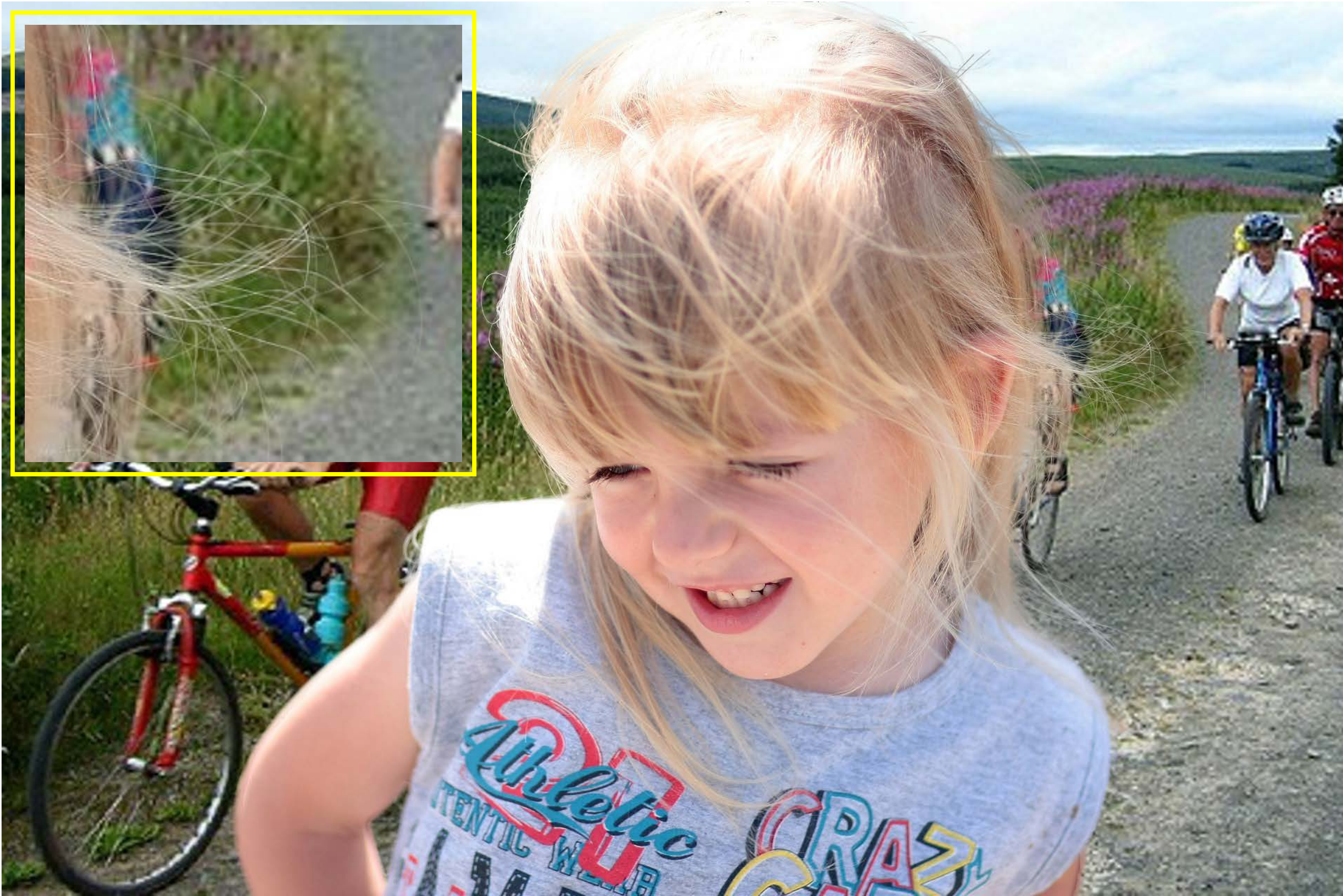}
\centerline{\small Input Image}
\label{img}
\end{minipage}}\hspace{-.15cm}\vspace{-.6cm}
\subfigure{
\begin{minipage}[t]{0.19\linewidth} 
\includegraphics[width=1\linewidth]{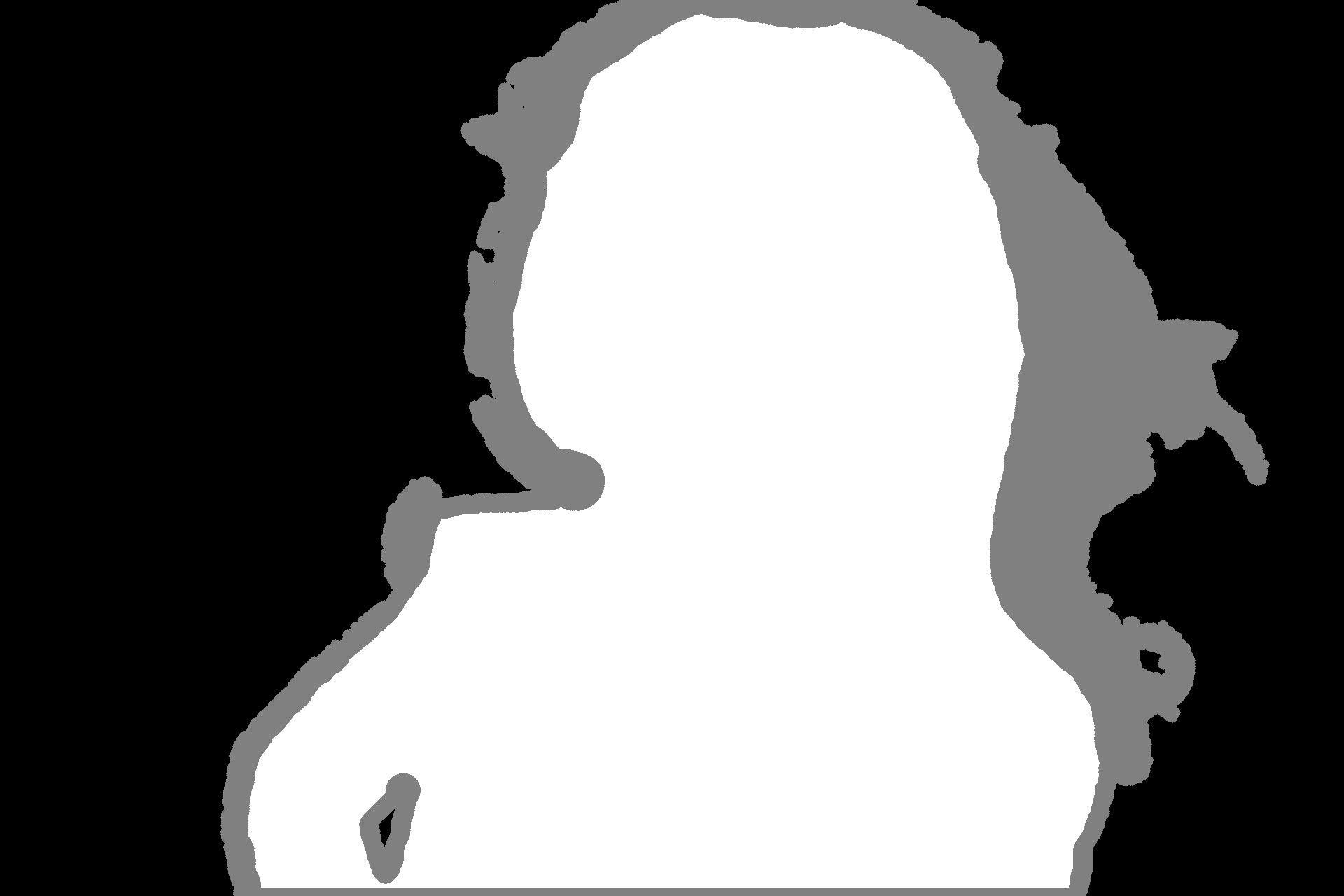}
\centerline{\small Trimap}
\label{img}   
\end{minipage}}\hspace{-.15cm}
\subfigure{
\begin{minipage}[t]{0.19\linewidth}   
 \includegraphics[width=1\linewidth]{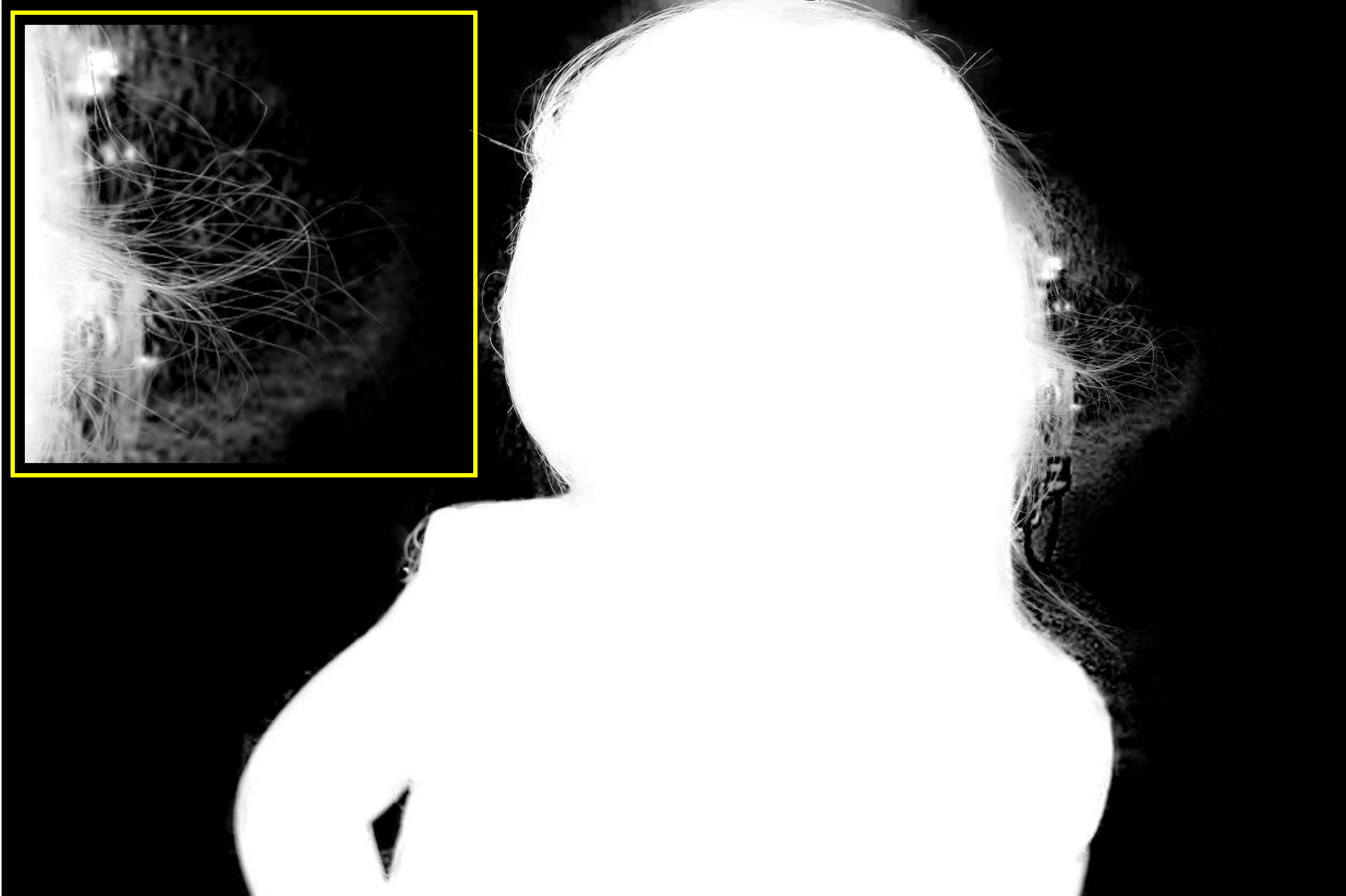}
 \centerline{\small Closed Form\cite{levin2008closed}}
 \label{img}    
 \end{minipage}}\hspace{-.15cm}
 \subfigure{
\begin{minipage}[t]{0.19\linewidth}   
 \includegraphics[width=1\linewidth]{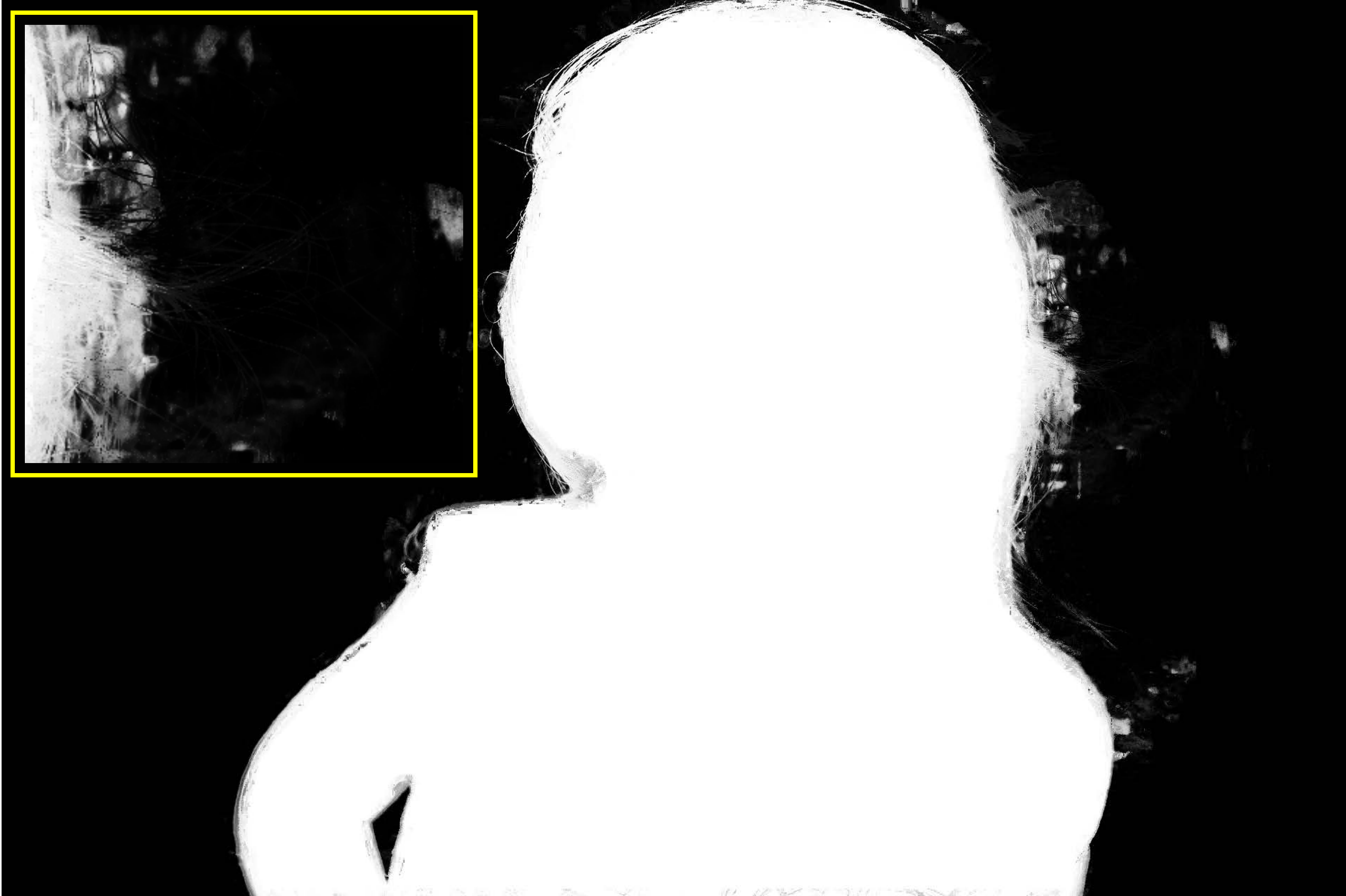}
 \centerline{\small KNN\cite{chen2013knn}}
 \label{img}    
 \end{minipage}}\hspace{-.15cm}
 \subfigure{
\begin{minipage}[t]{0.19\linewidth}   
 \includegraphics[width=1\linewidth]{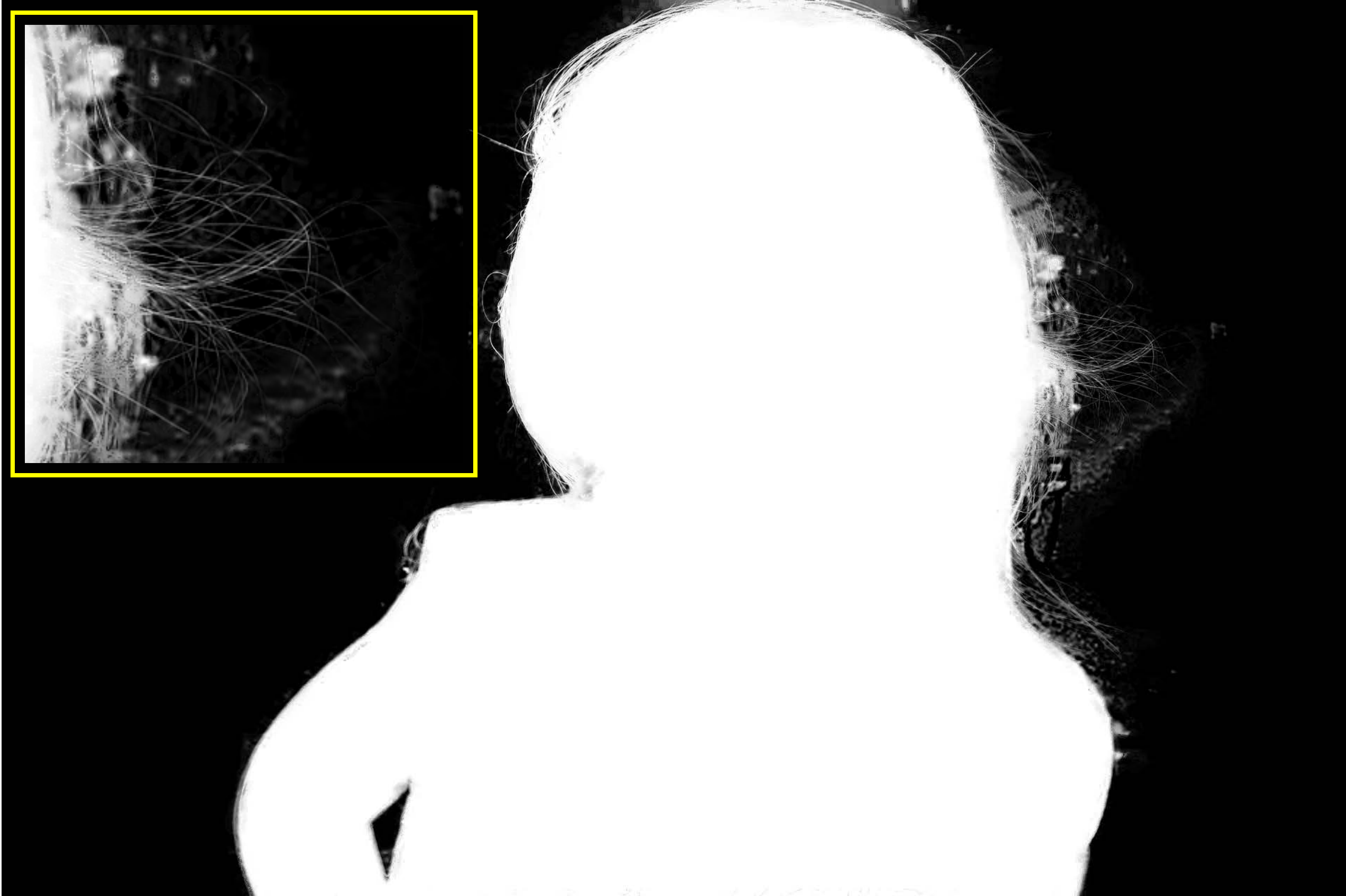}
 \centerline{\small DCNN\cite{cho2016natural}}
 \label{img}    
 \end{minipage}}\hspace{-.15cm}
\subfigure{
\begin{minipage}[t]{0.19\linewidth}   
 \includegraphics[width=1\linewidth]{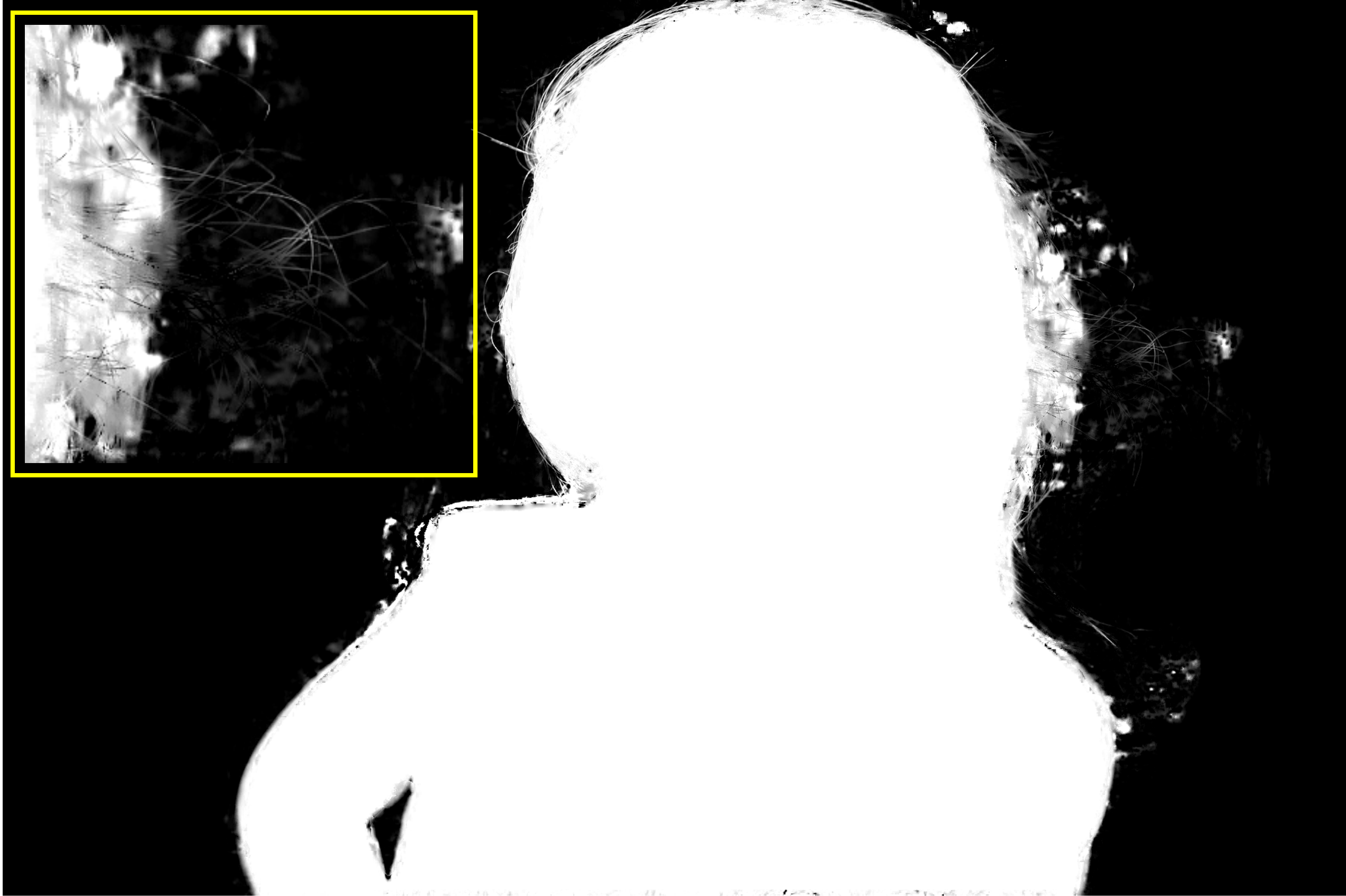} 
 \centerline{\small SM\cite{gastal2010shared}}
 \label{img}    
 \end{minipage}}\hspace{-.15cm}\vspace{-.6cm}
  \subfigure{
\begin{minipage}[t]{0.19\linewidth} 
 \includegraphics[width=1\linewidth]{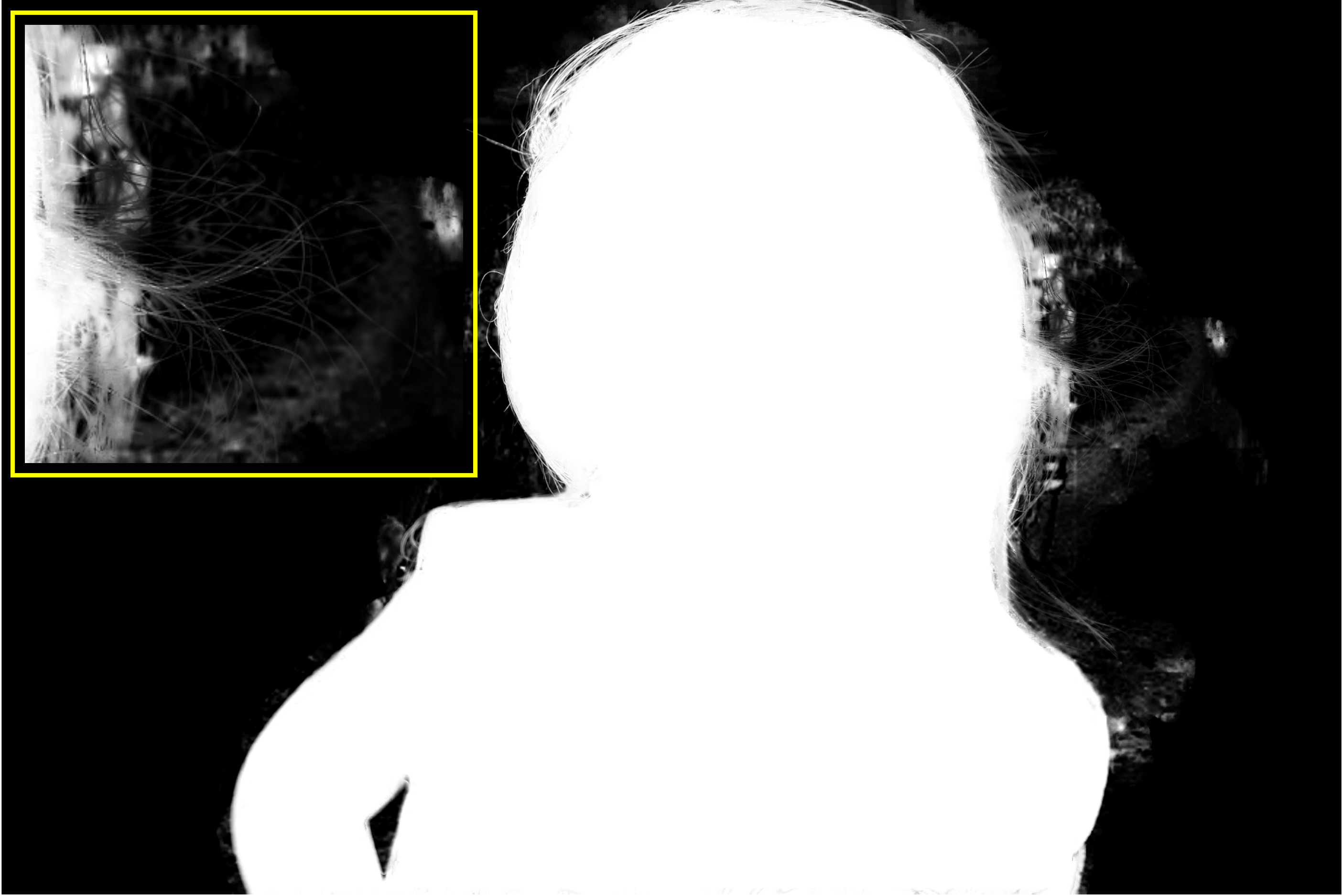} 
 \centerline{\small Information Flow\cite{aksoy2017designing}}
 \label{img}    
 \end{minipage}}\hspace{-.15cm}
 \subfigure{
\begin{minipage}[t]{0.19\linewidth}   
 \includegraphics[width=1\linewidth]{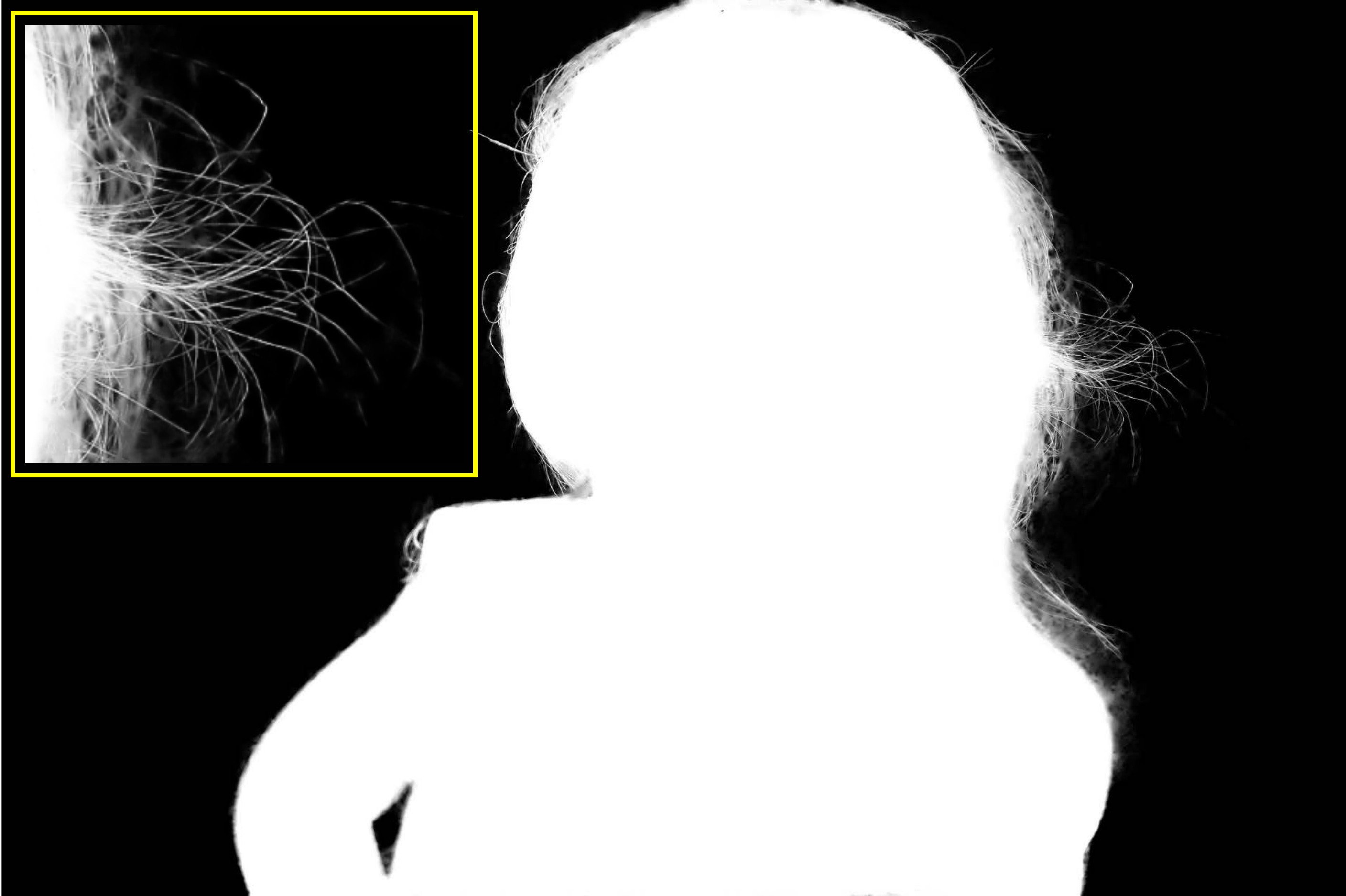} 
 \centerline{\small DIM\cite{xu2017deep}}
 \label{img}    
 \end{minipage}}\hspace{-.15cm}
 \subfigure{
\begin{minipage}[t]{0.19\linewidth}   
 \includegraphics[width=1\linewidth]{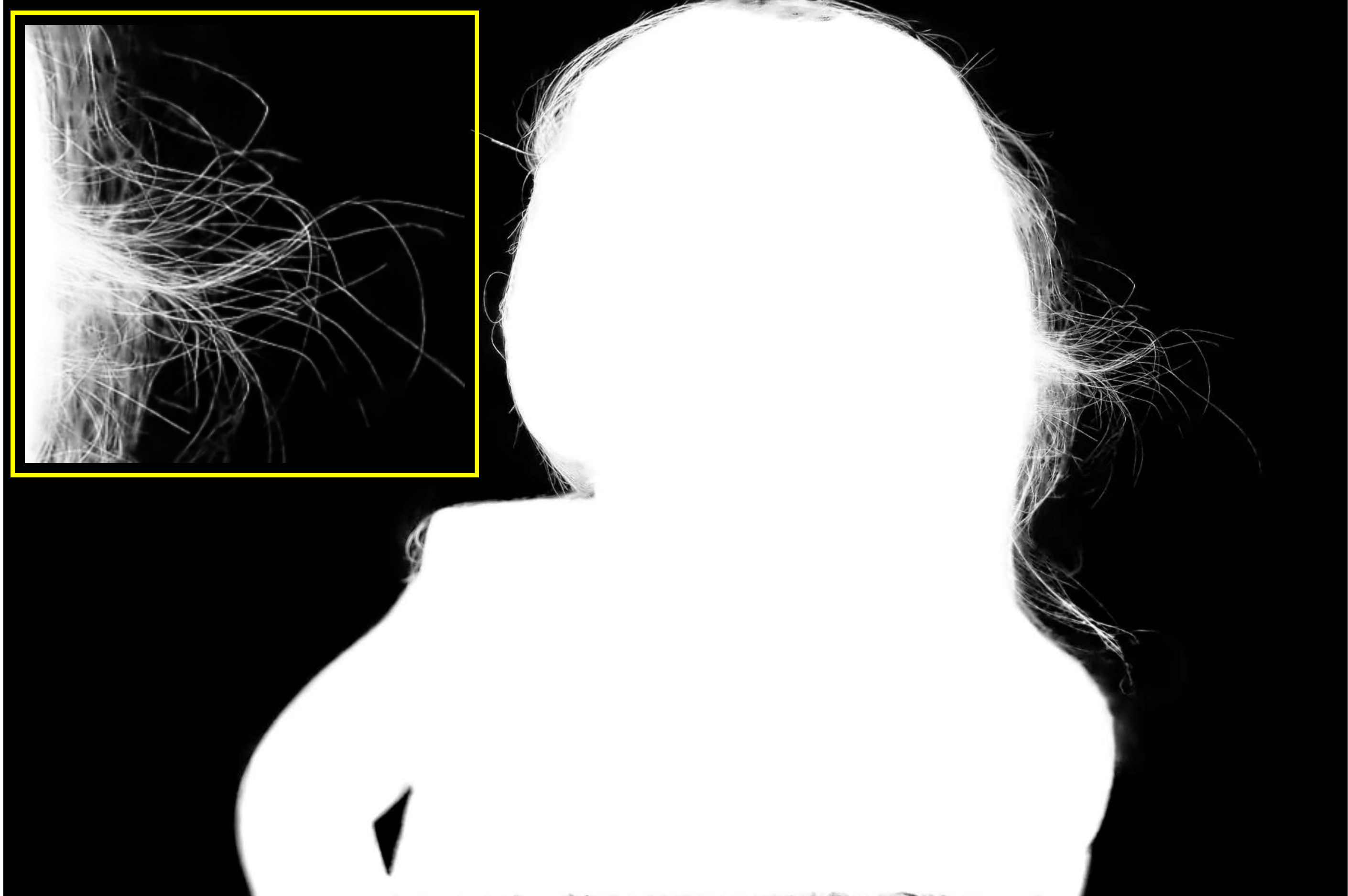} 
 \centerline{\small Ours}
 \label{img}    
 \end{minipage}}\hspace{-.15cm}
\subfigure{
\begin{minipage}[t]{0.19\linewidth} 
\includegraphics[width=1\linewidth]{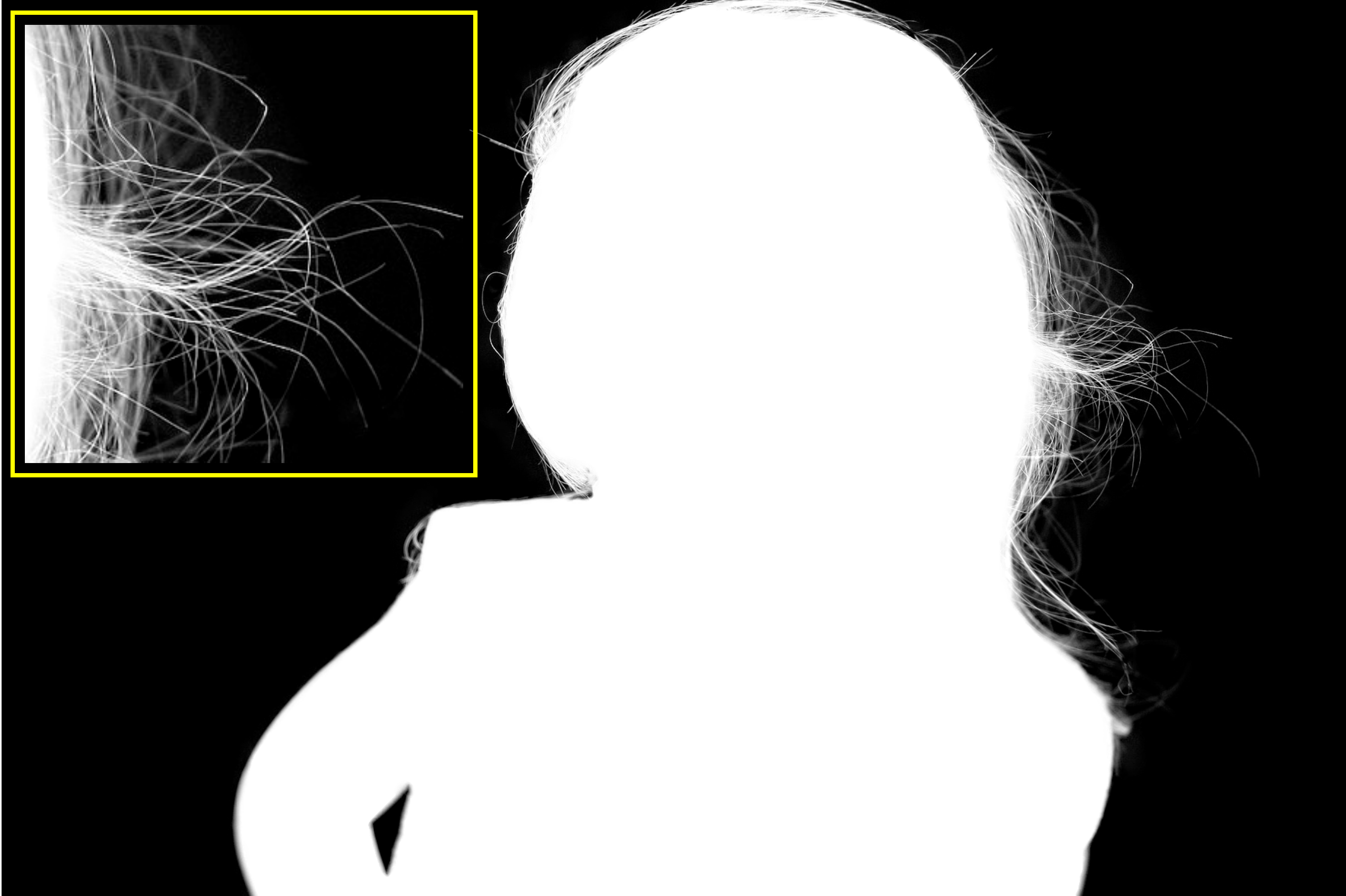}
\centerline{\small Ground Truth}
\label{img}
\end{minipage}}\hspace{-.15cm}

\subfigure{
\begin{minipage}[t]{0.19\linewidth} 
\includegraphics[width=1\linewidth]{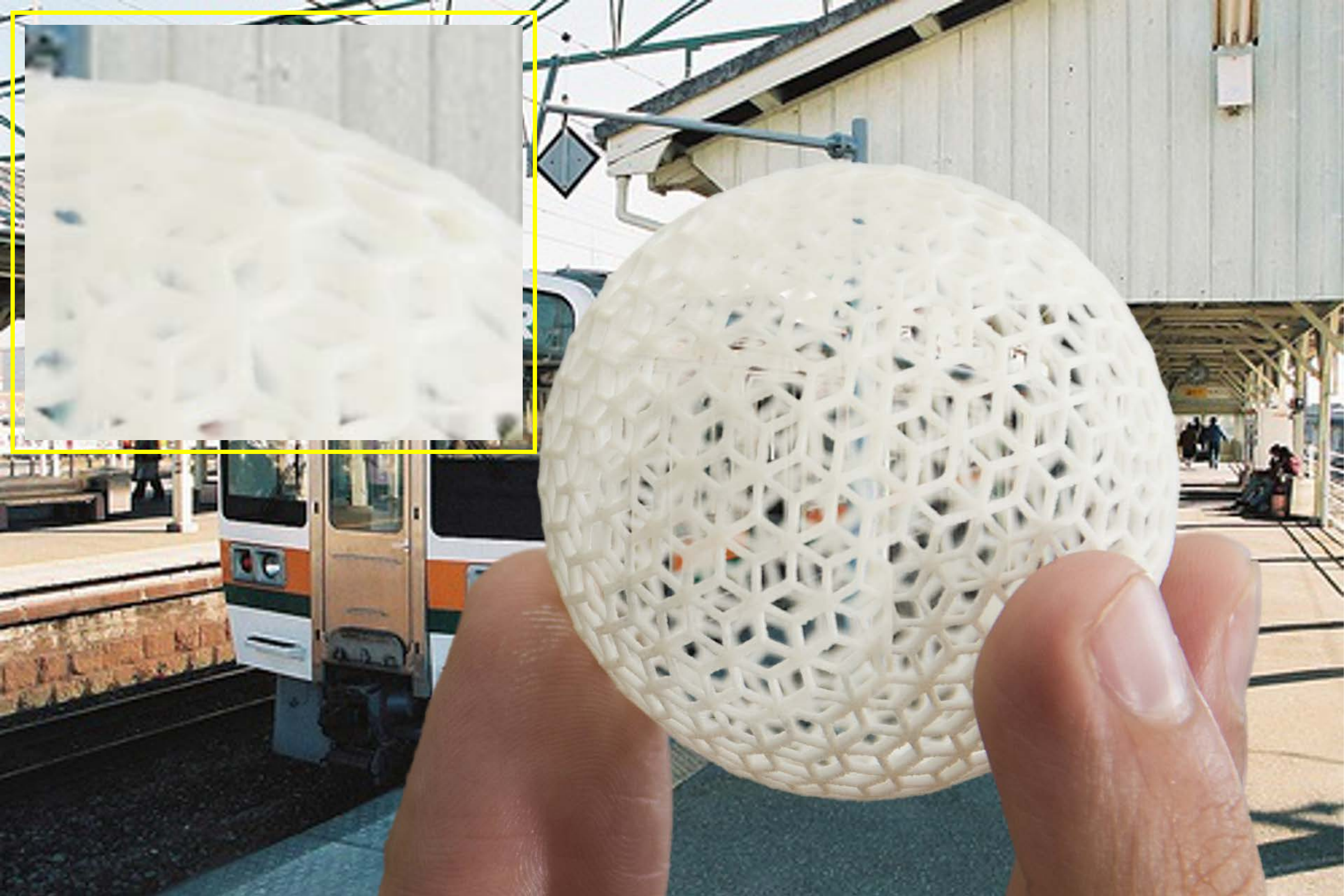}
\centerline{\small Input Image}
\label{img}
\end{minipage}}\hspace{-.15cm}\vspace{-.6cm}
\subfigure{
\begin{minipage}[t]{0.19\linewidth} 
\includegraphics[width=1\linewidth]{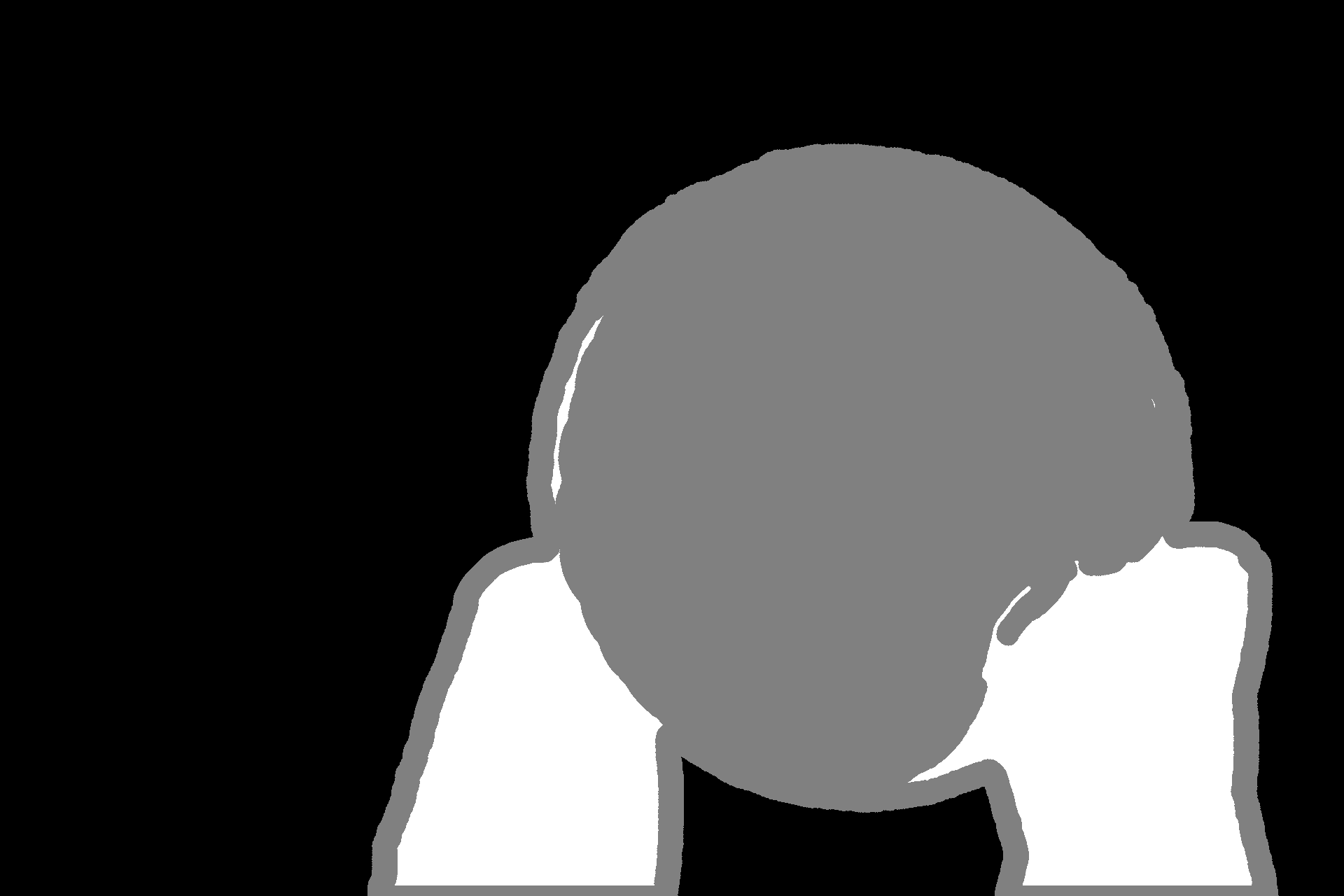}
\centerline{\small Trimap}
\label{img}   
\end{minipage}}\hspace{-.15cm}
\subfigure{
\begin{minipage}[t]{0.19\linewidth}   
 \includegraphics[width=1\linewidth]{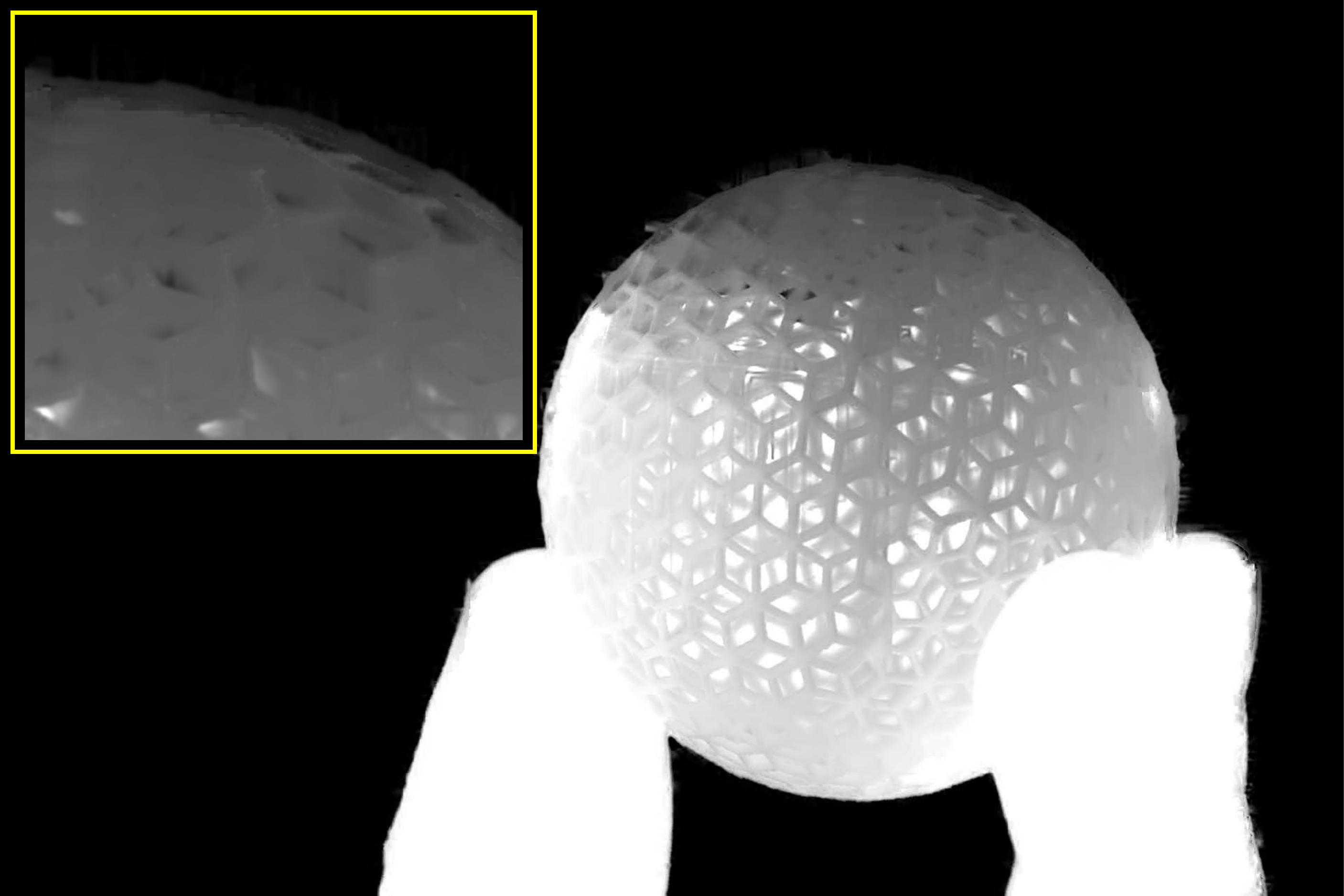}
 \centerline{\small Closed Form\cite{levin2008closed}}
 \label{img}    
 \end{minipage}}\hspace{-.15cm}
 \subfigure{
\begin{minipage}[t]{0.19\linewidth}   
 \includegraphics[width=1\linewidth]{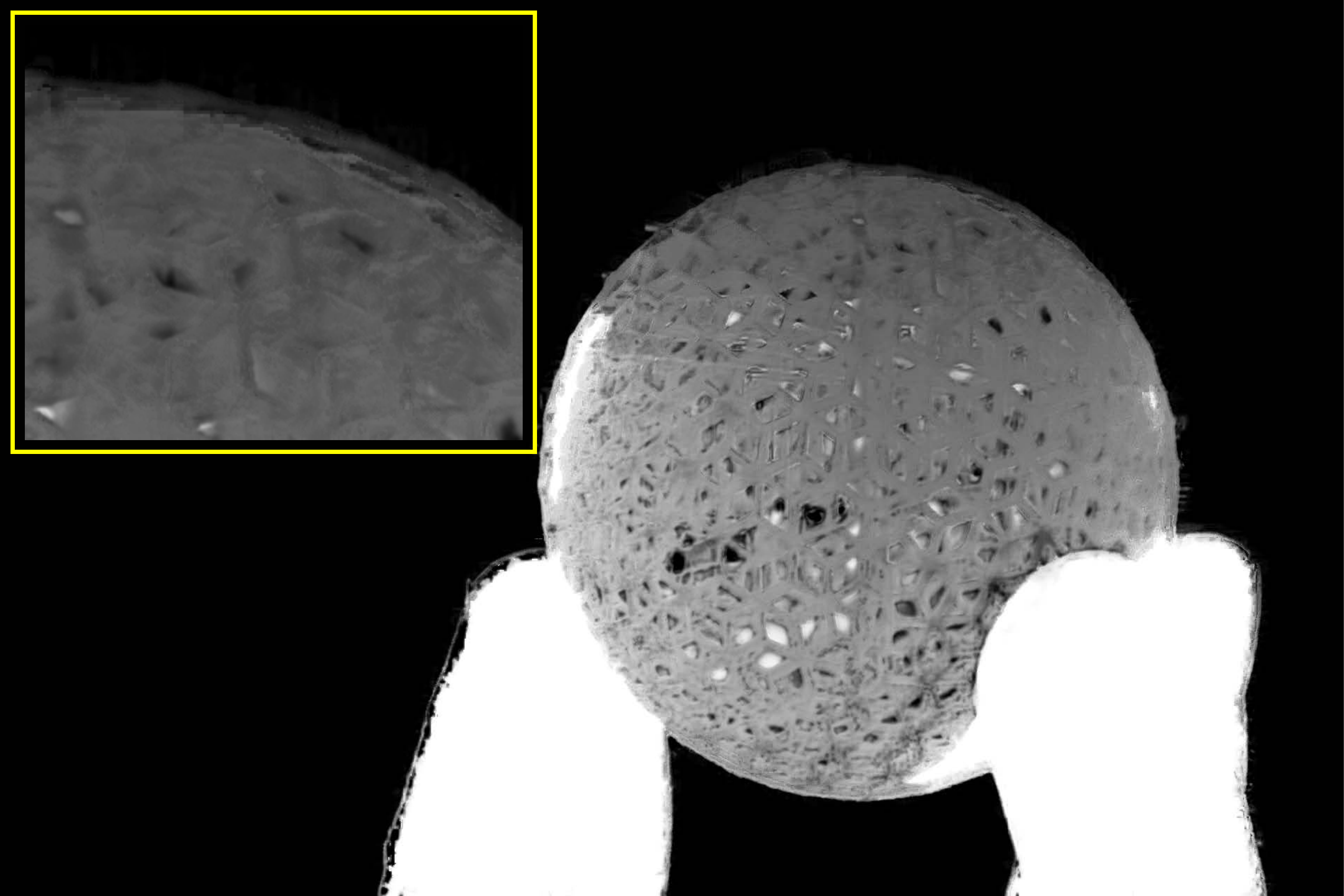}
 \centerline{\small KNN\cite{chen2013knn}}
 \label{img}    
 \end{minipage}}\hspace{-.15cm}
 \subfigure{
\begin{minipage}[t]{0.19\linewidth}   
 \includegraphics[width=1\linewidth]{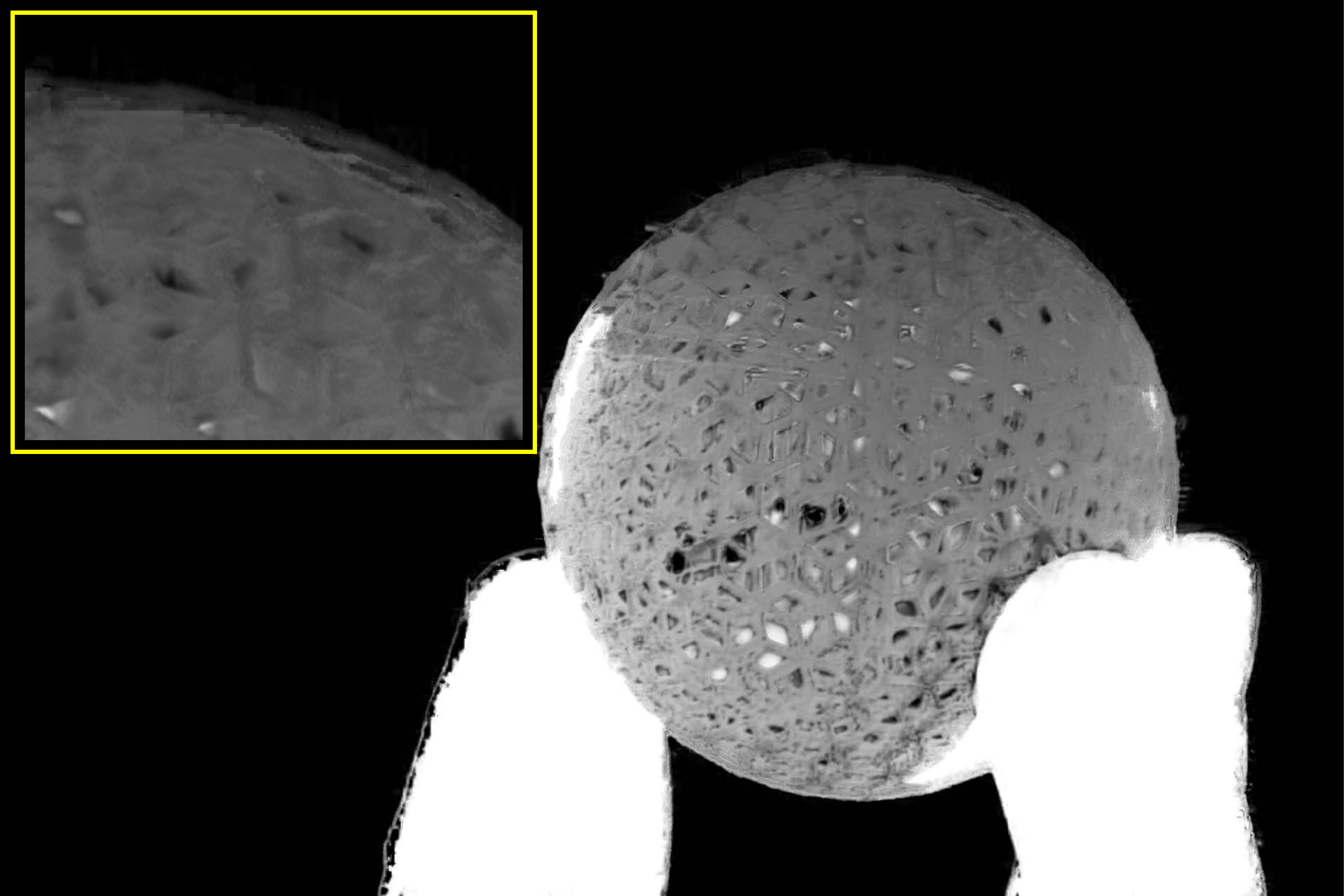}
 \centerline{\small DCNN\cite{cho2016natural}}
 \label{img}    
 \end{minipage}}\hspace{-.15cm}
\subfigure{
\begin{minipage}[t]{0.19\linewidth}   
 \includegraphics[width=1\linewidth]{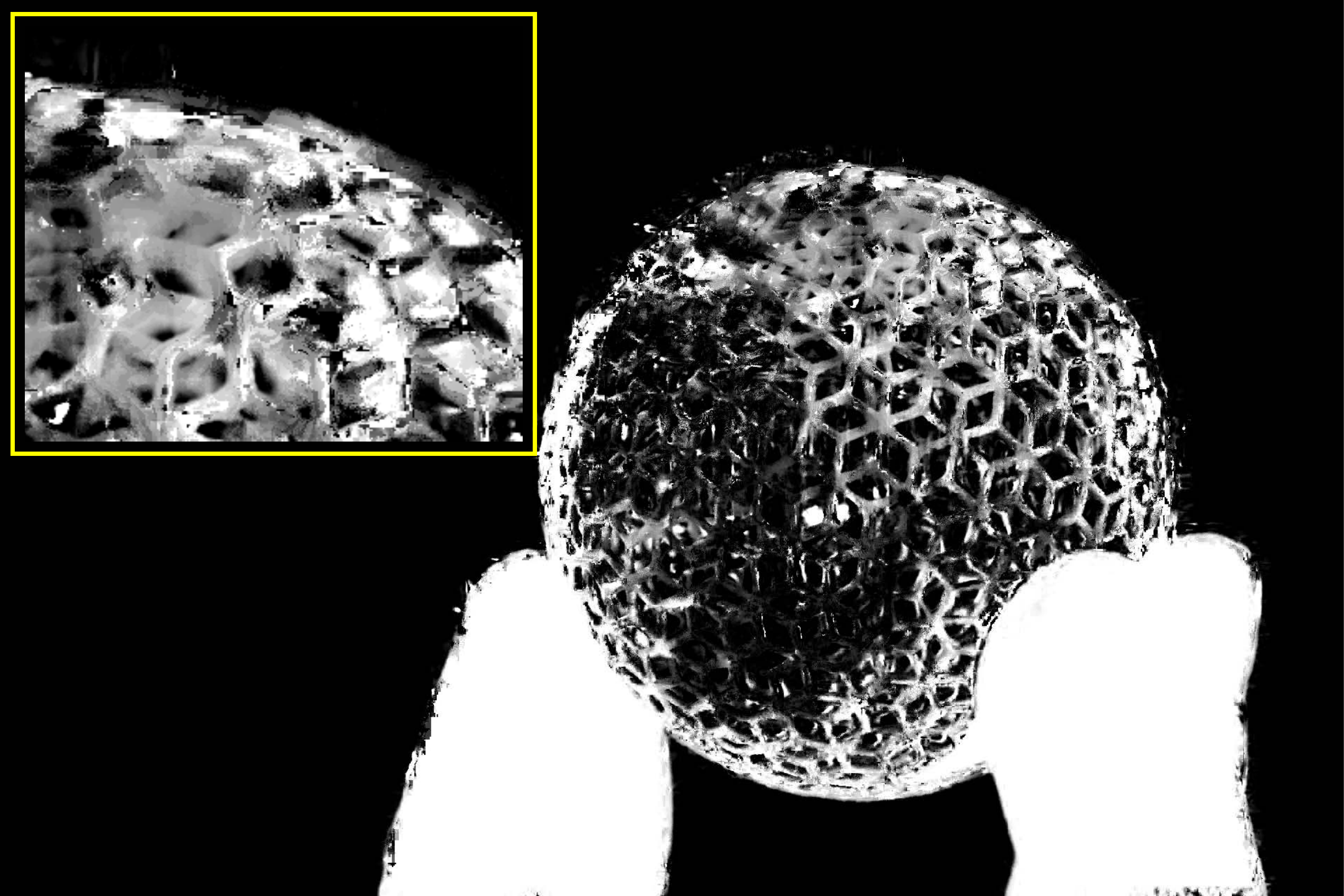} 
 \centerline{\small SM\cite{gastal2010shared}}
 \label{img}    
 \end{minipage}}\hspace{-.15cm}
  \subfigure{
\begin{minipage}[t]{0.19\linewidth} 
 \includegraphics[width=1\linewidth]{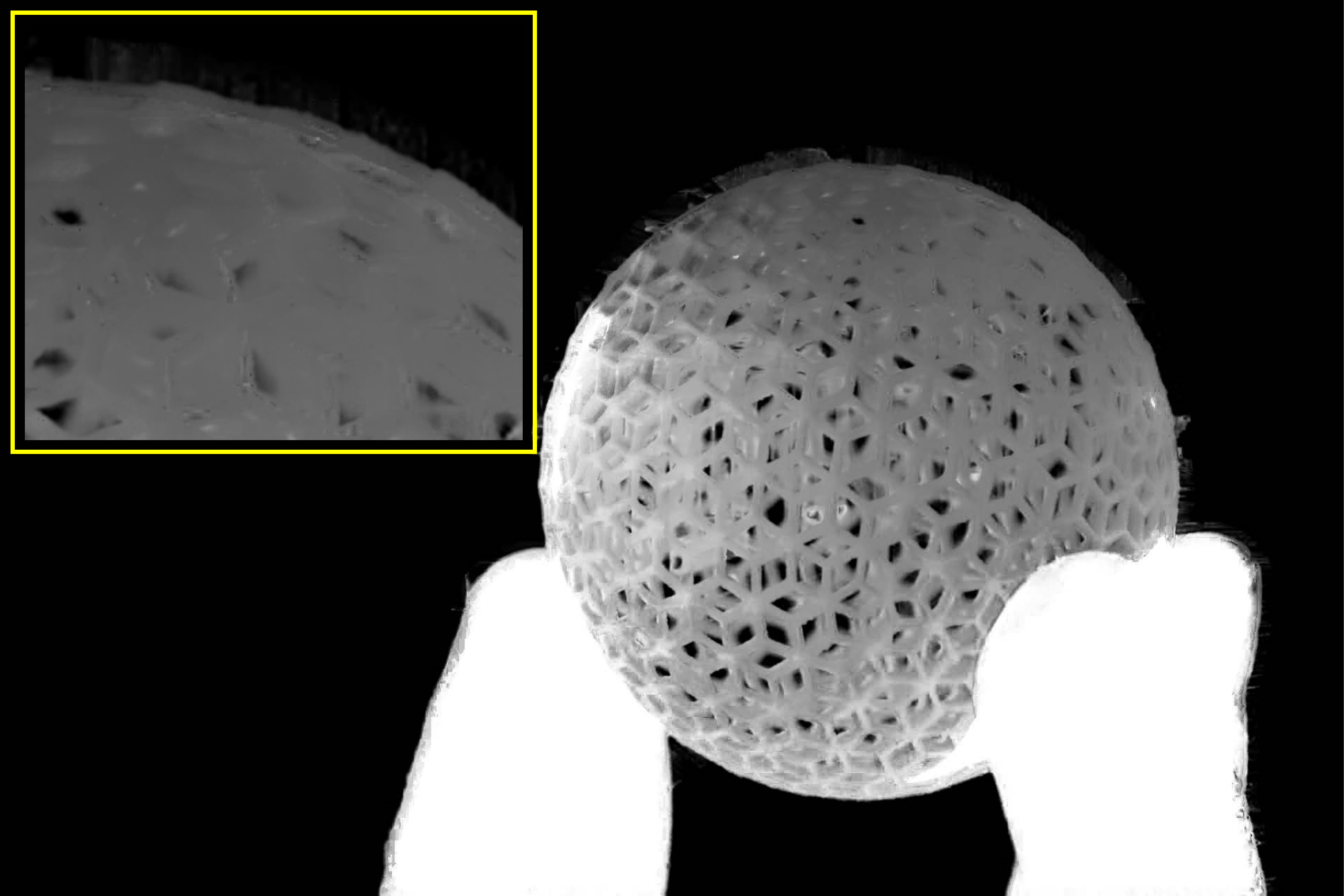} 
 \centerline{\small Information Flow\cite{aksoy2017designing}}
 \label{img}    
 \end{minipage}}\hspace{-.15cm}
 \subfigure{
\begin{minipage}[t]{0.19\linewidth}   
 \includegraphics[width=1\linewidth]{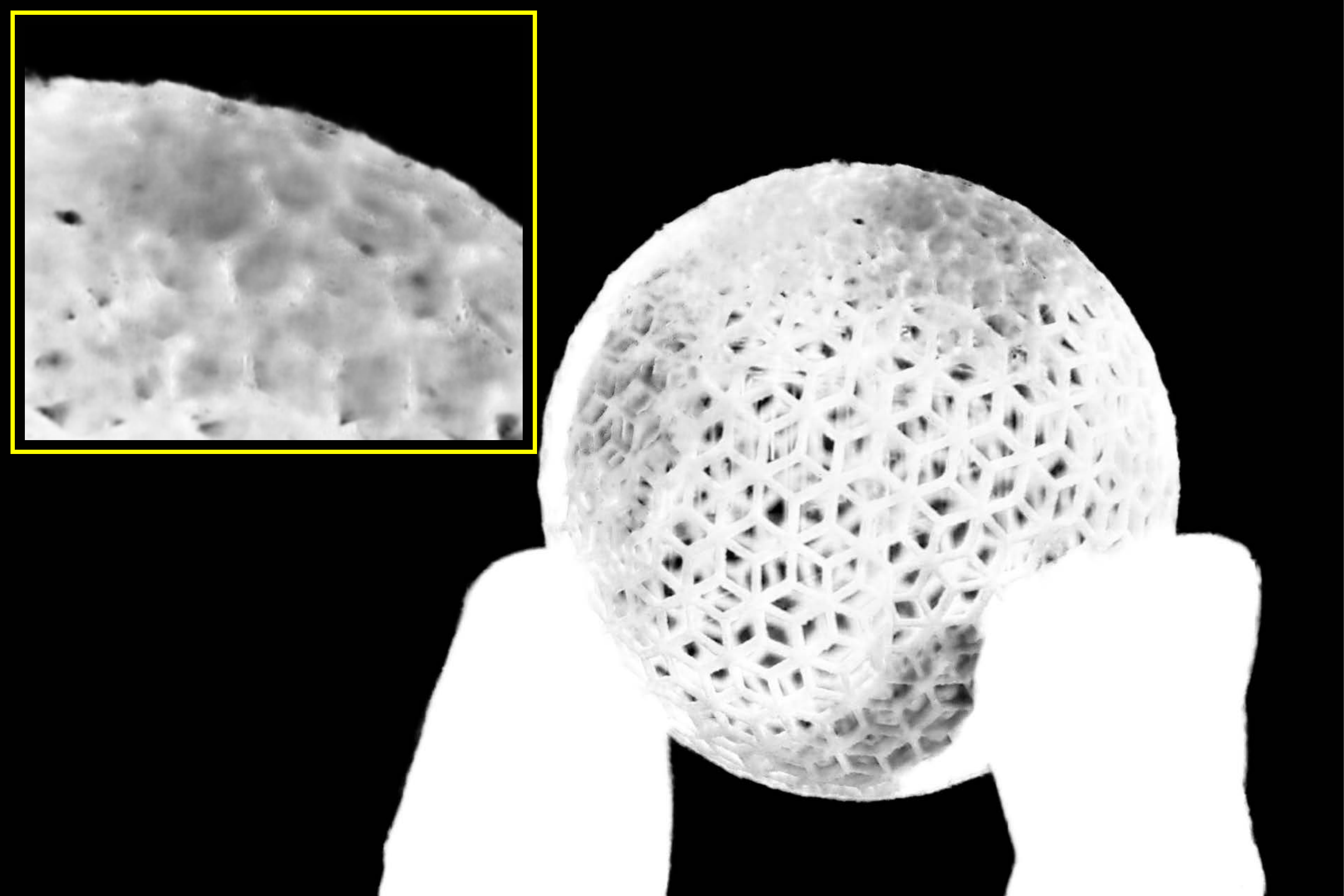} 
 \centerline{\small DIM\cite{xu2017deep}}
 \label{img}    
 \end{minipage}}\hspace{-.15cm}
 \subfigure{
\begin{minipage}[t]{0.19\linewidth}   
 \includegraphics[width=1\linewidth]{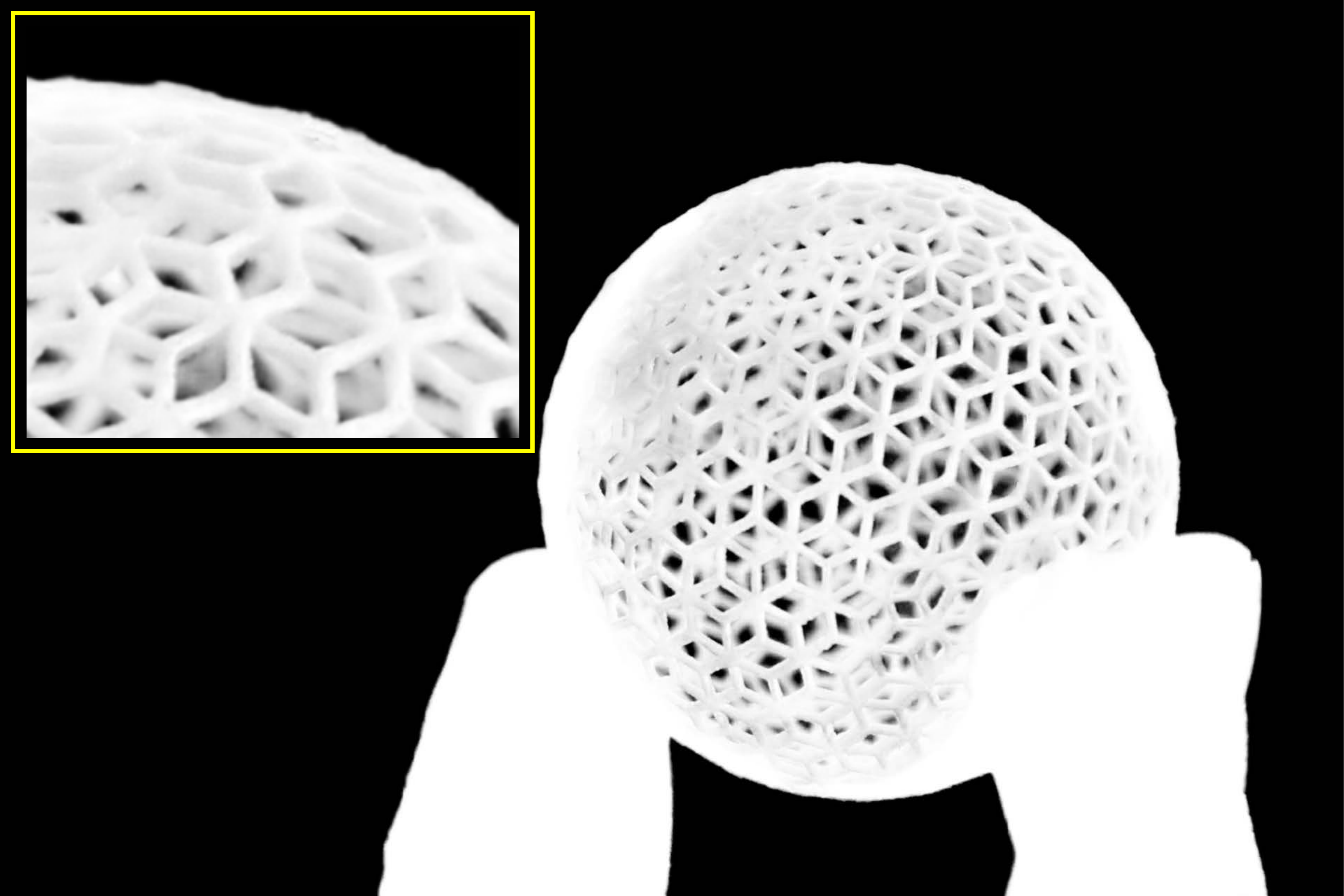} 
 \centerline{\small Ours}
 \label{img}    
 \end{minipage}}\hspace{-.15cm}
\subfigure{
\begin{minipage}[t]{0.19\linewidth} 
\includegraphics[width=1\linewidth]{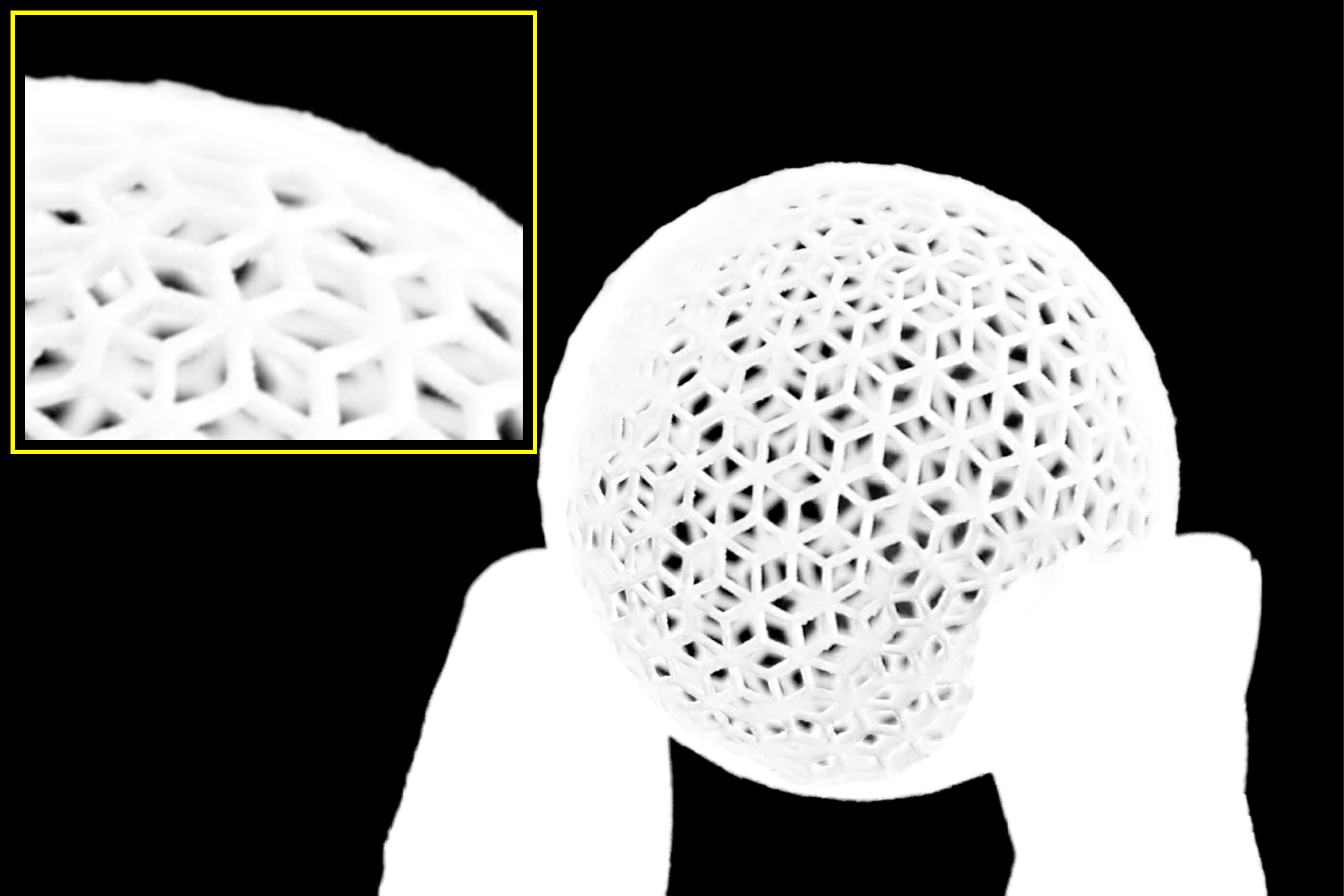}
\centerline{\small Ground Truth}
\label{img}   
\end{minipage}}
\vspace{-0.2cm}
\caption{Qualitative comparisons on the Adobe Composition-1k test set.}
\vspace{-0.1cm}
\label{fig:adobe_1k} 
\end{figure*}

As random backgrounds are selected to combine with each foreground object in the dataset, many images do not seem natural or realistic. Moreover, some particularly difficult images are presented in the data set, in which the foreground color is hard to be distinguished from the background. Two examples are shown in Fig. \ref{fig:adobe_1k}. It can be obviously seen that our results contain more vivid details and significantly less artifacts, compared to all other methods.

\setlength{\tabcolsep}{15pt}
\begin{table}
\begin{center}
\caption{Quantitative comparisons on the Composition-1k test set with other state-of-the-arts. The gradient loss is scaled by $10^3$. -PU stands for removing the Propagation Unit.}
\label{table:adobe}
\begin{tabular}{|c|c|c|c|}
\hline
Methods &Grad &SAD &MSE \\
\hline
CF \cite{levin2008closed}  &126.9& 168.1 & 0.091  \\
KNN \cite{chen2013knn}  &124.1& 175.4 & 0.103  \\
DCNN \cite{cho2016natural}  &115.1 & 161.4 & 0.087    \\
IF \cite{aksoy2017designing} & 38.0 & 52.4 & 0.030\\
AlphaGAN \cite{Lutz2018} & - &  - & 0.031 \\
DIM \cite{shen2016deep}  & 30.0 & 50.4 & 0.014  \\
Ours (-PU)  & 17.9 & 44.1 & 0.011  \\
Ours  & \textbf{16.8} & \textbf{41.7} & \textbf{0.010}  \\
\hline
\end{tabular}
\end{center}
\vspace{-0.9cm}
\end{table}
\setlength{\tabcolsep}{1.4pt}

\vspace{-0.2cm}
\section{Discussions}
\label{sec:discussions}
From the comparisons to state-of-the-art image matting models, it is obvious that our AdaMatting achieves superior performance both quantitatively and qualitatively. In this section, we conduct more experiments to further analyze the effectiveness of each designed technique, as well as measuring our performance on unseen real-world images.


\subsection{Comparison to the Two-Stage Method}
\label{sec:twostage}
Rather than training a single network in the multi-tasking manner, a more intuitive method is utilizing two cascaded networks, solving trimap adaptation and then image matting sequentially. We call such model Seq-AdaMatting, which does not share intermediate representations between the two tasks.

For fair comparisons, all components including sub-pixel convolutions, global convolutions, propagation unit and multi-task loss are employed for both models. The quantitative results on Adobe Composition-1k is listed in Table. \ref{tab:twostage}. It can be seen from the table that the original AdaMatting which utilizes the shared representations significantly outperforms the sequential version on all three metrics, despite the fact that the Seq-AdaMatting has much more parameters than the AdaMatting (since the Seq-AdaMatting have two different encoders). These results further prove that using shared representations, which contain rich semantic information learned from trimap adaptation, could effectively help to extract better alpha mattes.
\setlength{\tabcolsep}{10pt}
\begin{table}[!htbp]
\begin{center}
\caption{Quantitative comparisons to the two-stage sequential method. The gradient loss is scaled by $10^3$.}
\vspace{0.3cm}
\label{tab:twostage}
\begin{tabular}{|c|c|c|c|}
\hline
 Method & Grad & SAD & MSE \\
\hline
AdaMatting - w/o PU & \textbf{17.93} & \textbf{44.10} & \textbf{0.0114} \\
Seq-AdaMatting  & 23.97 & 46.36 & 0.0129 \\
\hline
\end{tabular}
\end{center}
\vspace{-0.6cm}
\end{table}

\begin{figure*}[!htp]
\vspace{-0.2cm}
\begin{center}
\subfigure{
\begin{minipage}[t]{0.16\linewidth} 
\includegraphics[width=1\linewidth]{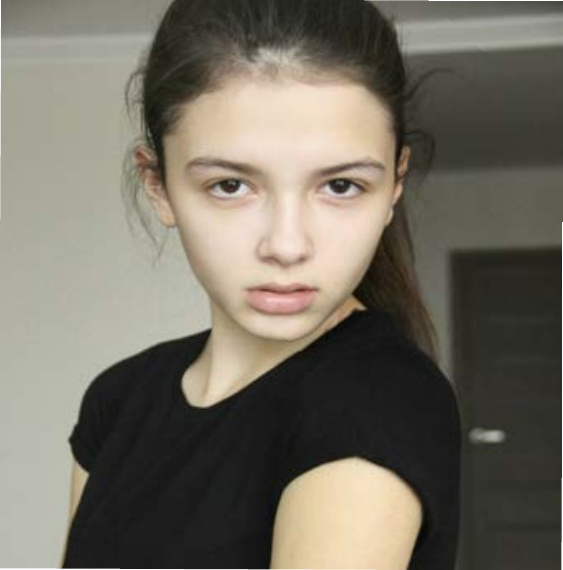}
\centerline{\small Input Image}
\label{img}
\end{minipage}}\hspace{-.15cm}\vspace{-.1cm}
\subfigure{
\begin{minipage}[t]{0.16\linewidth}   
 \includegraphics[width=1\linewidth]{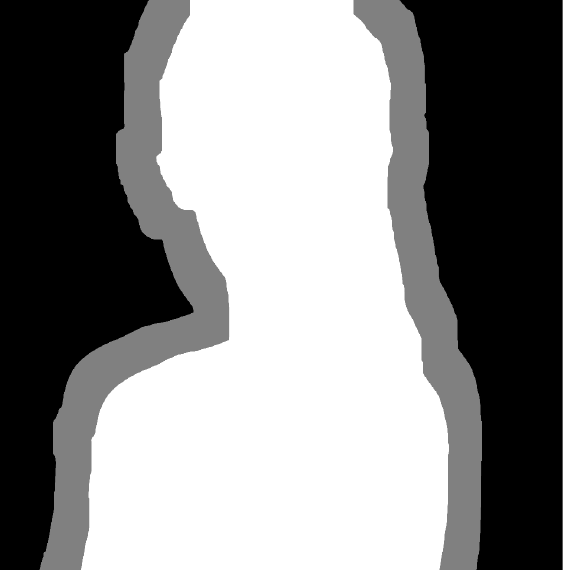}
 \centerline{\small Input Trimap }
 \label{img}    
 \end{minipage}}\hspace{-.15cm}\vspace{-.1cm}
  \subfigure{
\begin{minipage}[t]{0.16\linewidth} 
 \includegraphics[width=1\linewidth]{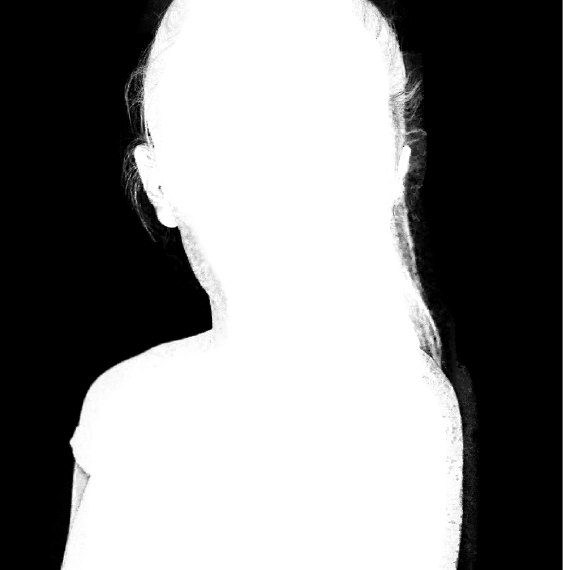}
 \centerline{\small IF \cite{aksoy2017designing}}
 \label{img}    
 \end{minipage}}\hspace{-.15cm}\vspace{-.1cm}
 \subfigure{
\begin{minipage}[t]{0.16\linewidth}   
 \includegraphics[width=1\linewidth]{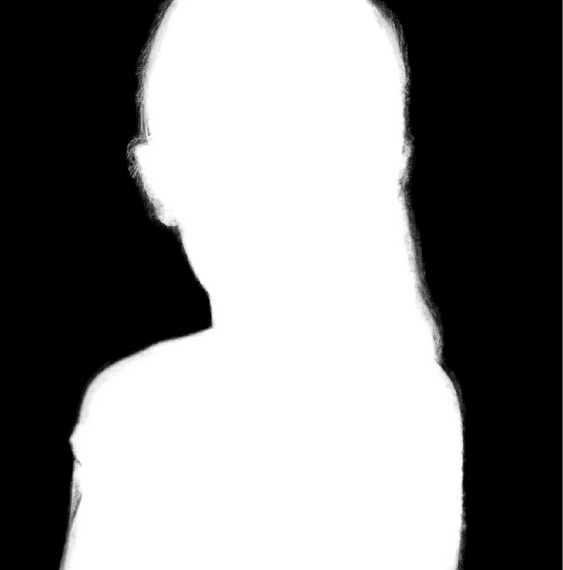}
 \centerline{\small DIM \cite{xu2017deep}}
 \label{img}    
 \end{minipage}}\hspace{-.15cm}\vspace{-.1cm}
  \subfigure{
\begin{minipage}[t]{0.16\linewidth}   
 \includegraphics[width=1\linewidth]{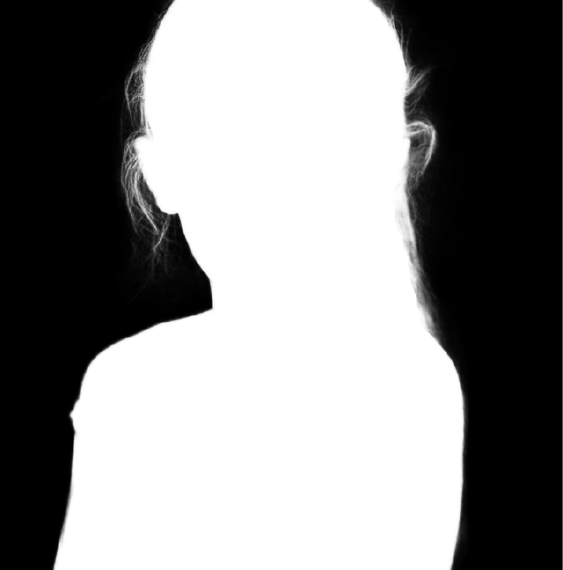} 
 \centerline{\small Ours \cite{lutz2018alphagan}}
 \label{img}    
 \end{minipage}}\hspace{-.15cm}\vspace{-.1cm}
 \subfigure{
\begin{minipage}[t]{0.16\linewidth}   
 \includegraphics[width=1\linewidth]{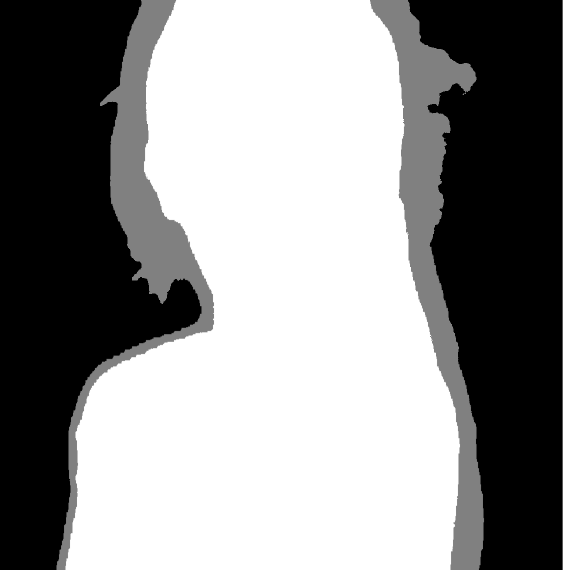} 
 \centerline{\small Our Adapted Trimap}
 \label{img}    
 \end{minipage}}\hspace{-.15cm}\vspace{-.1cm}
\end{center}
\vspace{-0.1cm}
\caption{Evaluation results on a real-world image. The input trimap is generated by portrait segmentation followed by boundary eroding.}
\label{fig:real_image}
\vspace{-0.7cm}
\end{figure*}
\subsection{Accuracy of Trimap Adaptation}

 We have provided visualization examples of  adapted trimap in paper and supp. For quantitative results, please refer to Tab.~\ref{tab:Accuracy}.

\setlength{\tabcolsep}{8pt}
\begin{table}\vspace{-0.35cm}
\begin{center}
\caption{Perf. of trimap adaptation (Acc, mIoU) and image matting (Grad) on Adobe's testset\cite{xu2017deep}. ``D-n'': Using n-dilation of GT alpha as input trimap. ``Adobe'': Using testset trimap as input. }\vspace{-0.3cm}
\begin{tabular}{|c|c|c|c|}
\hline
Method / Trimap Type & Acc (\%) &mIoU &Grad \\
\hline
CSS Matting~\cite{shahrian2013improving} / D-100  &84.9  &59.0  &480.99 \\
CSS Matting~\cite{shahrian2013improving} / D-10  & 92.3 &77.8  &129.8 \\
CSS Matting~\cite{shahrian2013improving} / Adobe &90.3 & 77.2 & 116.27 \\
 \hline 
AdaMatting / D-100   &94.7 & 80.7 & 17.68  \\
AdaMatting / D-10  & 96.7 & 84.2 & 17.06  \\
AdaMatting / Adobe & 96.5 & 83.6 & 16.89 \\
\hline
\end{tabular}
\label{tab:Accuracy}
\end{center}
\vspace{-1.15cm}
\end{table}
\setlength{\tabcolsep}{1.0pt}

\subsection{Effectiveness of the Structural Semantics}
\begin{figure}
\centering
\subfigure{
 \begin{minipage}[t]{0.45\linewidth}   
 \includegraphics[width=1\linewidth]{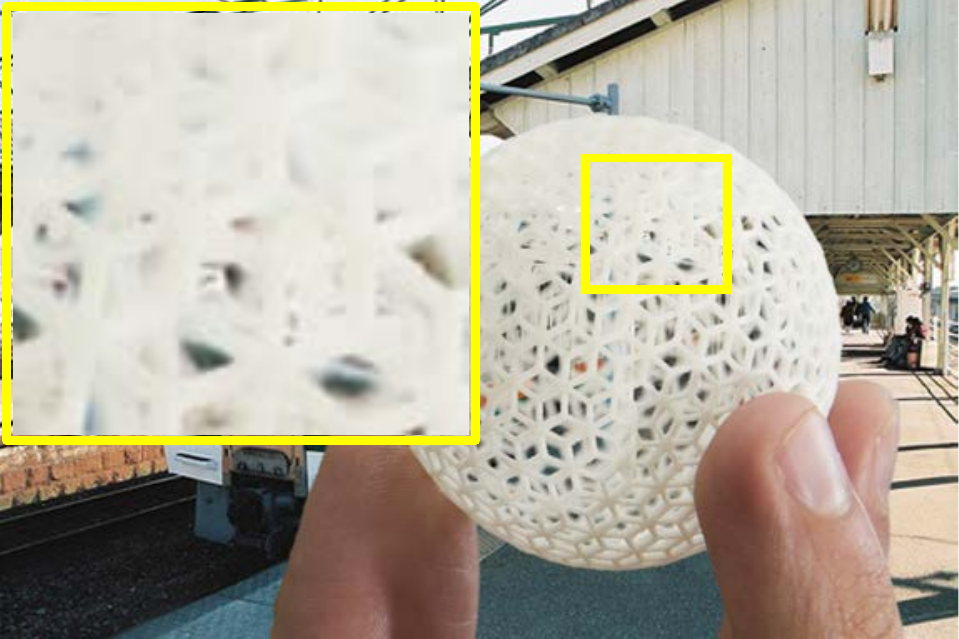}
 \centerline{\small Input Image}\vspace{-0.15cm}
 \end{minipage}}
 \subfigure{
 \begin{minipage}[t]{0.45\linewidth}   
 \includegraphics[width=1\linewidth]{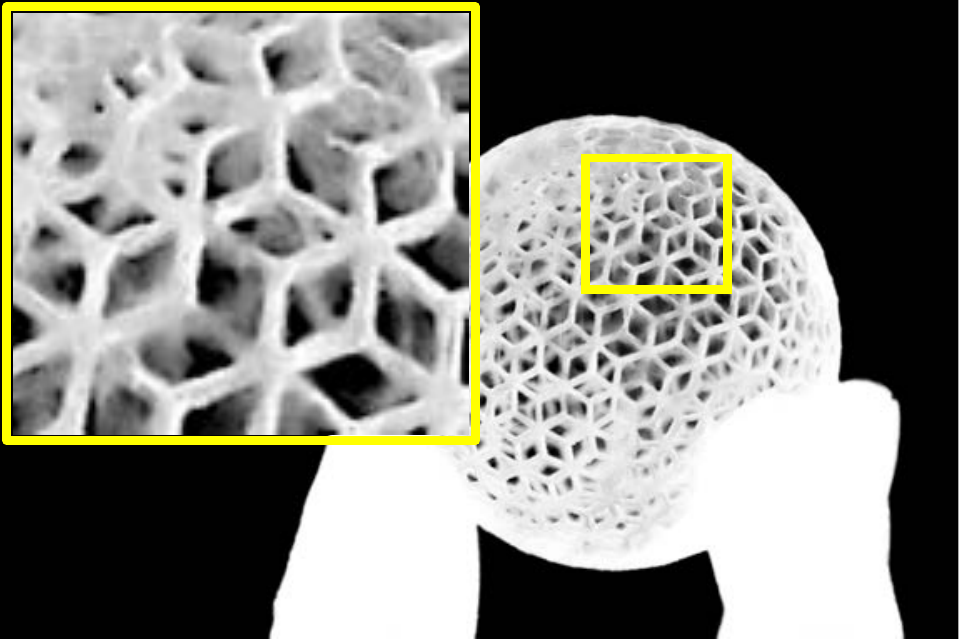}
 \centerline{\small AdaMatting with $\mathcal{L}_\alpha$ Only}\vspace{-0.15cm}
 \end{minipage}}
 \subfigure{
 \begin{minipage}[t]{0.45\linewidth}   
 \includegraphics[width=1\linewidth]{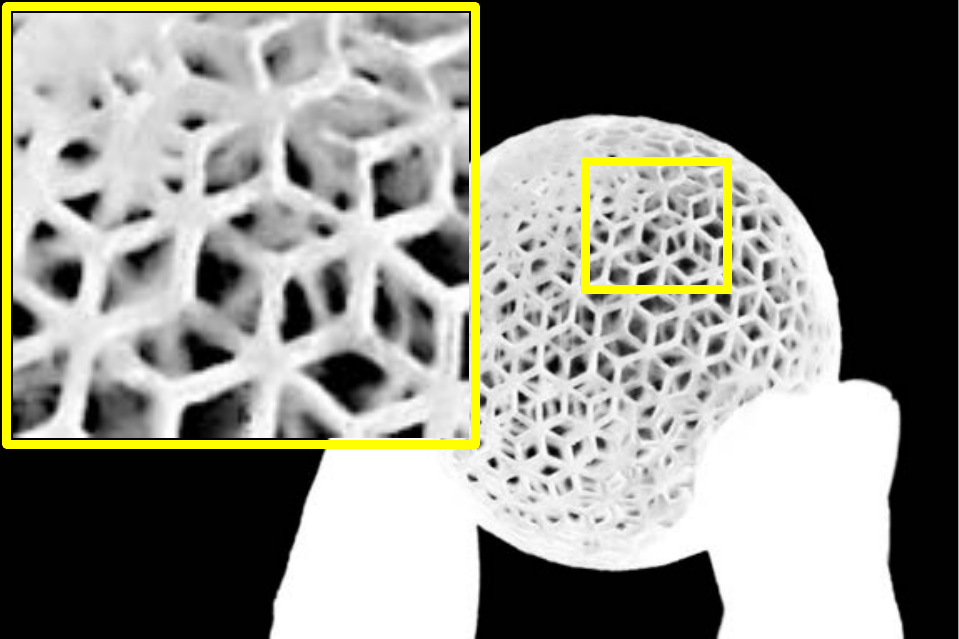}
 \centerline{\small Seq-AdaMatting}
 \end{minipage}}
\subfigure{
\begin{minipage}[t]{0.45\linewidth}
\includegraphics[width=1\linewidth]{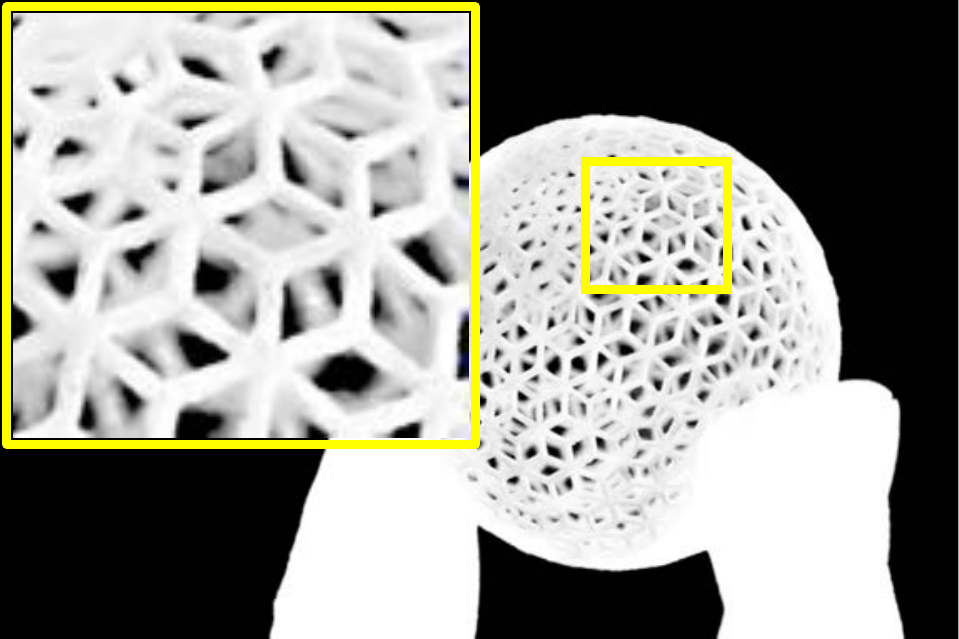}
 \centerline{\small Proposed AdaMatting}
\end{minipage}}
\caption{Comparisons of our model trained with and without structural semantic features from trimap adaptation. Obviously the one with these features (the last alpha mattes) could capture the overall structure, and yield a more accurate result.}   
\label{fig:semantics}
\vspace{-0.6cm}
\end{figure}

To further prove the effectiveness of leading in structural semantics, we designed an experiment to compare our proposed AdaMatting (with multi-task loss and shared representations) with those trained without the help of semantic information. The first is the model trained only with the alpha estimation loss $\mathcal{L}_\alpha$ (\textit{i.e.} $\sigma=1.0$ in Eq. \ref{eq:linearloss}), not involving high-level semantics learned from trimap adaptation. The second is the two-stage method mentioned in Section \ref{sec:twostage}, which does not share representations between the two tasks. Thus the alpha estimation step is not guided by the semantic features from trimap adaptation as well.

We take a closer look to the second image in Fig. \ref{fig:adobe_1k}, since the ball is highly structured, contains sophisticated patterns on a large scale. Furthermore, the color resemblance between the foreground and background add to the overall difficulty for image matting. Thus solving this image would definitely need the global perception to the object shape and structure. The alpha matte results of the three models are shown in Fig. \ref{fig:semantics}. It can be obviously seen that with the structural semantics learned from trimap adaptation, the proposed model could precisely capture the overall shape of the foreground object, leading to accurate matting results without loss of details. The other models without high-level features fail to perceive the global structure of the object, resulting in deficiencies in the alpha mattes.

\subsection{Analysis on the Multi-Task Loss}
To further analysis the impact of the multi-task loss. We carry out experiments on the Adobe Composition-1k testset, comparing two kinds of losses: the deployed dynamically weighted loss (Eq. \ref{eq:mtloss}) and naive linearly combined loss:

\vspace{-0.35cm}
\begin{equation}
\label{eq:linearloss}
\mathcal{L}_{naive} = (1-\sigma) \mathcal{L}_T + \sigma \mathcal{L}_\alpha,
\vspace{-0.1cm}\end{equation}
where $\sigma$ stands for a pre-defined fixed weight.

\begin{figure}
\begin{center}
\includegraphics[width=1.0\linewidth]{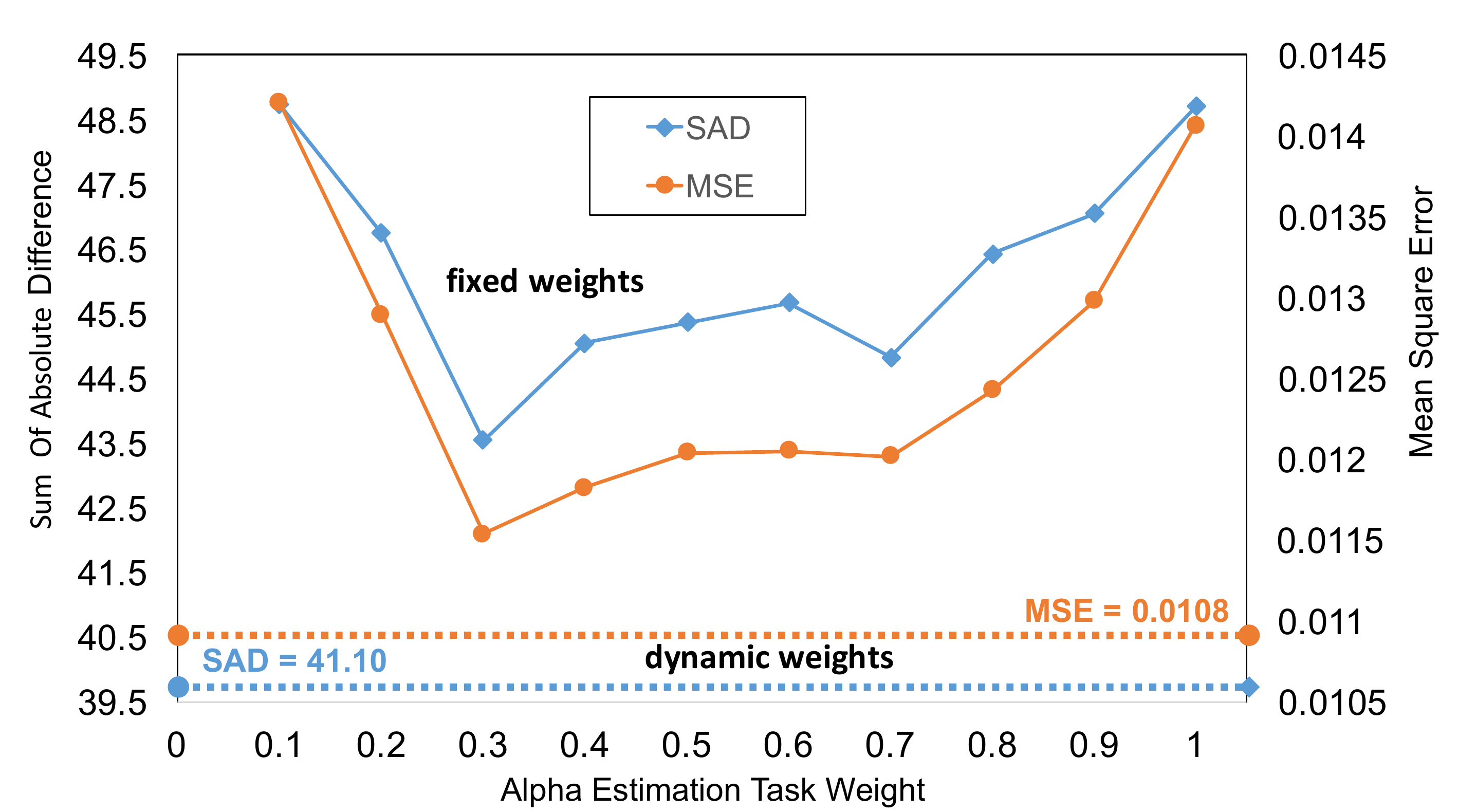}
\end{center}
\vspace{-0.2cm}
\caption{Results on the Adobe Composition-1k test set with different weighted loss functions. The dotted line below represents the result with dynamically weighted loss, and the poly lines above represent the result with linearly combined fixed-weight loss.}
\vspace{-0.6cm}
\label{fig:weight}
\end{figure}

Note that in the extreme case of $\sigma=1.0$, the method degenerates to the one step regression of alpha used by previous work\cite{xu2017deep, cho2016natural}. We train the AdaMatting under same settings except for the loss function. The resulted model performance with respect to weight $\sigma$ is shown in Fig. \ref{fig:weight}.

It can be observed that $\sigma=1.0$ leads to significantly inferior performance, which verifies the importance of solving trimap adaptation explicitly. Properly adjusting the weight between the classification branch and the regression branch improves performance. However, the dynamically weighted loss leads to markedly better results compared to all other losses. 


\subsection{Real-World Image Matting}
Because of the additional robustness provided by trimap adaptation, our model could generate accurate alpha mattes even when the input trimaps contain minor errors. The robustness is particularly useful when performing real-world image matting. One of the results is shown in Fig. \ref{fig:real_image}. As observed, our AdaMatting produces much more meticulous details compared to other methods. Furthermore, because of the task of trimap adaptation, our model is capable of correcting the input trimaps, yielding accurate alpha values even at the improperly labelled regions.

\section{Conclusions}
In this paper, we proposed a disentangled view of image matting, where the task can be divided into two sub-tasks: trimap adaptation and alpha estimation. From this point of view, the AdaMatting is proposed to solve both sub-tasks jointly utilizing the multi-task loss. By explicitly separating the two sub-tasks and optimizing them according to different objectives, the model can greatly benefit from the shared representations, which contains both rich semantic and photometric information. Extensive experiments demonstrate additional structural awareness and trimap fault-tolerance of the AdaMatting. Furthermore, the proposed method shows superior performance on two widely used datasets, both qualitatively and quantitatively, establishing a new state-of-the-art for image matting.


{\small
\bibliographystyle{ieee_fullname}
\bibliography{egbib}
}

\end{document}